\documentclass[prodmode,acmtkdd]{acmprep} 
\usepackage{url}
\usepackage{dcolumn}
\usepackage{hyperref}
\usepackage[capitalise]{cleveref}
\usepackage{graphicx}
\usepackage{morefloats}
\usepackage{rotating}

\usepackage[ruled]{algorithm2e}
\renewcommand{\algorithmcfname}{ALGORITHM}
\SetAlFnt{\small}
\SetAlCapFnt{\small}
\SetAlCapNameFnt{\small}
\SetAlCapHSkip{0pt}
\IncMargin{-\parindent}

\acmArticle{AD}

\begin{document}

\markboth{Emmott, Das, Dietterich, Fern, Wong}{A Meta-Analysis of the Anomaly Detection Problem}

\title{A Meta-Analysis of the Anomaly Detection Problem}
\author{ANDREW EMMOTT
\affil{Oregon State University}
SHUBHOMOY DAS
\affil{Oregon State University}
THOMAS DIETTERICH
\affil{Oregon State University}
ALAN FERN
\affil{Oregon State University}
WENG-KEEN WONG
\affil{Oregon State University}}

\begin{abstract}
This article provides a thorough meta-analysis of the anomaly detection problem. To accomplish this we first identify approaches to benchmarking anomaly detection algorithms across the literature and produce a large corpus of anomaly detection benchmarks that vary in their construction across several dimensions we deem important to real-world applications: (a) point difficulty, (b) relative frequency of anomalies, (c) clusteredness of anomalies, and (d) relevance of features. We apply a representative set of anomaly detection algorithms to this corpus, yielding a very large collection of experimental results. We analyze these results to understand many phenomena observed in previous work. First we observe the effects of experimental design on experimental results. Second, results are evaluated with two metrics, ROC Area Under the Curve and Average Precision. We employ statistical hypothesis testing to demonstrate the value (or lack thereof) of our benchmarks. We then offer several approaches to summarizing our experimental results, drawing several conclusions about the impact of our methodology as well as the strengths and weaknesses of some algorithms. Last, we compare results against a trivial solution as an alternate means of normalizing the reported performance of algorithms. The intended contributions of this article are many; in addition to providing a large publicly-available corpus of anomaly detection benchmarks, we provide an ontology for describing anomaly detection contexts, a methodology for controlling various aspects of benchmark creation, guidelines for future experimental design and a discussion of the many potential pitfalls of trying to measure success in this field.
\end{abstract}





\begin{bottomstuff}
This research was supported in part by the Defense Advanced Research Projects Agency (DARPA) under Contract W911NF-11-C-0088 and in part by the Future of Life Institute (futureoflife.org) FLI-RFP-AI1 program under grant number 2015-145014.  Any opinions, findings and conclusions or recommendations expressed in this material are those of the authors and do not necessarily reflect the views of the Future of Life Institute, DARPA, the Army Research Office, or the US government.
\end{bottomstuff}

\maketitle

\newpage

\section{Introduction}\label{sec:intro}

Anomaly detection is an important inference task with applications across many different domains including identifying novel threats in computer security \cite{lane1997sequence,portnoy2001intrusion,Lazarevic:2003,Lazarevic:2007}, discovering novel astronomical phenomena \cite{wltdg-gsdeud-2013}, detecting broken environmental sensors \cite{Dereszynski2011}, identifying machine component failures \cite{Xue:06,Zhang:08,Alzghoul:2011}, and finding cancer cells in normal tissue \cite{Polat:05,greensmith:06}. Despite the importance of the task, the field of statistical anomaly detection lacks a standard methodology for understanding and evaluating proposed algorithms.  Most published experiments evaluate their algorithms via application-specific case studies or ad hoc synthetic datasets.  There are very few realistic, publicly-available benchmark datasets. There are two consequences of this. First, it is very difficult to compare different algorithms to assess progress in the field. Second, it is difficult to understand the various factors or dimensions of anomaly detection problems that influence the performance of anomaly detection algorithms. This makes it difficult for experiments to guide research in algorithm development.

Building on previous work \cite{odd2013}, we aim to provide a thorough meta-analysis of the anomaly detection problem. In this study we apply what we hope is a representative set of anomaly detection algorithms and in so doing we are able to provide a robust comparison of these algorithms against each other, but this article is not intended as a survey of algorithms nor even a definitive comparison of the algorithms herein. Instead we aim to highlight the common pitfalls associated with evaluating success in this field and the difficulty in measuring progress in the field.

As in our previous work we develop and test a standardized evaluation methodology for statistical anomaly detection. There are several important differences in methodology and evaluation of results than in \cite{odd2013}. For those familiar with our previous work, you can find a description of these differences in \cref{appendix:prev}, but rather than assume familiarity with our previous work, the body of this article will describe our methodology in full.

The remainder of this article is organized as follows. We first refine our definition of the anomaly detection problem to make it clear what domains we are and are not considering in this study. Next we review and assess existing approaches to the evaluation of anomaly detection methods. Based on this, in \cref{sec:require} we identify a set of requirements for experimental methodology and identify four problem dimensions we believe are relevant to anomaly detection applications. In \cref{sec:method} we present our benchmarking methodology and provide detailed procedures for meeting our requirements and controlling our problem dimensions. In \cref{sec:algo} we present the field of anomaly detection algorithms used in this study with parameterization details noted in \cref{appendix:parameters}. In \cref{sec:hypo} we identify our evaluation metrics and describe the statistical hypothesis tests we apply to each algorithm's result on each benchmark. A summary of these hypothesis tests and a discussion of these results follows in the same section. In \cref{sec:basic} we present our most straightforward findings, comparing the impact of many measurable quantities against their respective control groups. Later in that same section we offer several alternate views of our results, intended to both illuminate certain properties of some algorithms as well as highlight the difficulty of comparing algorithms in this field. In \cref{sec:discuss} we provide a global discussion of the findings presented in this paper and provide recommendations for how these findings should impact future work in the field.

\subsection{The Task of Anomaly Detection}

We study the following unsupervised anomaly detection setting. We are given a collection of $N$ data points $x_1, \ldots, x_N$, each a $d$-dimensional real-valued vector.  These data points are a mixture of ``nominal'' points and ``anomalous'' points. However, none of the points are labeled. The goal is to identify the anomalous points. 

An anomaly detection algorithm takes as input the $N$ data points and produces as output a real-valued anomaly score for each point such that points with higher scores are believed to be more anomalous. Natural metrics of quality for anomaly predictions are the area under the ROC curve (AUC) and the Average Precision (AP; also known as the area under the precision-recall curve). In some applications, we are only interested in the top $K$ highest-ranked points, in which case, natural metrics are the precision and recall at $K$. In other applications, we might choose a threshold and classify all points whose anomaly score exceeds the threshold as anomalies and all other points as nominal. In such settings, common metrics are precision, recall, and F1 (the harmonic mean of precision and recall). Accuracy or error rate are typically not very useful, because in most applications the anomalies constitute a very small fraction of the data (e.g., from 0.01\% to 1\%). In this paper, we consider only the AUC and AP metrics.

When confronted with an anomaly detection problem, there are two main algorithmic approaches. One is to model both the normal and the anomalous points. This can succeed if we have a good understanding (or sufficient labeled training data) for both kinds of data points. However, it is in the nature of anomaly detection problems that we usually lack a good understanding of the process that is generating the anomalous points.  In computer security \cite{zhang06,denning1987}, the adversaries are constantly changing their attack strategies, a fact that invalidates supervised learning studies such as those based on the KDD 99 Challenge Cup \cite{kdd99}. In astronomy, the goal is to discover surprising and unexpected phenomena.  For machine failure, there seems to be a Murphy's Law that machines will always find new ways to fail. Cancer is difficult to model, because it is not a single disease, but rather a whole constellation of different (and poorly-understood) mechanisms \cite{aggarwal2009}.

Instead of modeling the processes generating the anomalies, the second approach (and the one that we study) is to assume the problem is unsupervised and detect the anomalies by relying on statistical signals. Specifically, we look for statistical outliers and hope that those outliers are anomalies. We will refer to this assumption as the ``outliers-as-targets'' assumption. It is the predominant assumption in the literature.  It was nicely-articulated by Hawkins \cite{Hawkins1980}:
\begin{quote}
``An outlier is an observation which deviates so much from the other observations as to arouse suspicions that it was generated by a different mechanism.''
\end{quote}

From a methodological perspective, this means that when constructing benchmark datasets for anomaly detection, we should not just sample from a probability distribution with heavy tails. Instead, we should generate data by combining two different generating processes so that we can assess the relationship between being an outlier and being an anomaly.

\subsection{Existing Experimental Methodology}

In anomaly detection research, three kinds of data have been employed to analyze and evaluate anomaly detection algorithms. First, there are datasets drawn from specific application problems (e.g., \cite{Xue:06,Lazarevic:2003}). Second, there are synthetic datasets \cite{mulcross}. Third, there are datasets constructed by taking an existing supervised classification problem and treating one or more of the classes as the anomalies. 

Application-specific datasets are very useful. They can help understand and evaluate the algorithm refinements needed to achieve high performance in a particular application. However, often the datasets are not publicly available because of privacy or security considerations (e.g., \cite{senator2013a}). 

Synthetic datasets \cite{Liu:08,Liu:10,mulcross} permit the systematic manipulation of some properties (e.g., the relative frequency of the anomalies, the distinctiveness of the anomalies, etc).  However, decades of experience in machine learning have shown that real data sets are much more complex and idiosyncratic than synthetic data, which undermines the validity of this approach \cite{mahoney2003,mchugh2000}.

Finally, the repurposing of supervised classification data sets has the desirable property that the different classes are the result of different generating processes and the data retain the idiosyncrasies of the real application (e.g., \cite{Liu:08}). However, most studies have treated the datasets ``as is'' without trying to manipulate properties of the data. One exception is the work of Kim and Scott \cite{kim:08} who sub-sampled the anomaly class to reduce the relative frequency of the anomalies. Another interesting case is the work of Das, et al., \cite{das2008}, who generated anomalies by permuting features among a small subset of the data.  In a few cases, supervised regression datasets have been repurposed by treating the data points with the most extreme values as anomalies \cite{cortez2009}.

We propose to combine the idea of repurposing supervised learning datasets with the idea of systematically varying properties of the data.  In this paper, we show how to take existing supervised learning datasets and manipulate the relative frequency of the anomalies, the degree of difficulty of individual anomaly points, the degree to which the anomalies are clustered (versus scattered), and the degree to which the features are relevant versus irrelevant for the task.  Before describing our techniques, we first collect a list of the requirements that anomaly detection benchmarks should satisfy. 

\section{Requirements for Anomaly Detection Experiments}\label{sec:require}

As discussed above, although anomaly detection algorithms work by searching for statistical outliers, the goal is to identify points that are generated by a process that is distinct from the process generating the ``normal'' points. This distinction leads to the first two requirements for benchmark datasets.

\vspace*{0.1in}
\noindent{\bf Requirement 1: Normal data points should be drawn from a real-world generating process.}

\vspace*{0.1in}
\noindent{\bf Requirement 2: Anomalous data points should also be drawn
  from a real-world process, but one that is distinct from
  the process generating the normal points.}  The anomalous points
should not just be points in the tails of the ``normal''
distribution. See, for example, Glasser and Lindauer's synthetic
anomaly generator \cite{glasser2013}.

\vspace*{0.1in}
\noindent{\bf Requirement 3: Many benchmark datasets are needed.} If we
employ only a small number of datasets, we risk developing algorithms
that only work on those problems. More important, presenting results on many benchmarks at a time makes for more robust and reliable reported results. While the corpus of benchmarks presented in this study may be well in excess of what is needed for a reliable experiment, \cref{sec:dense} will illustrate the potential consequences of using too few benchmarks.
\vspace*{0.1in}
\noindent{\bf Requirement 4: Benchmark datasets should be characterized
in terms of well-defined and meaningful problem dimensions.} Applications of anomaly detection often face different challenges across domains. Experiments in the literature sometimes describe such challenges and propose strategies for addressing them, such as in ~\cite{Liu:10}. It is of practical value to real-world applications that experiments acknowledge which domain-specific challenges they might be addressing.

There is currently no established set of problem dimensions for anomaly detection. We have identified four dimensions that we believe are important, but we consider this only the beginning of the conversation. We only introduce the concepts here; how we attempt to measure these problem dimension is explained in \cref{sec:method}.

{\it Point difficulty:} The outliers-as-targets assumption breaks down as the target points become harder to distinguish from the normal points. One aspect of applying anomaly detection in adversarial settings (e.g., intrusion detection or insider threat detection) is that adversaries try to blend in with the distribution of normal points. We propose \emph{point difficulty} as a measure of the similarity of the anomalous data points to the normal ones.  When the targets are not confined to extreme outliers, or when the extreme outliers are not anomalies, the anomalies of interest will be confused with normal points or with uninteresting outliers.

This phenomenon has also been referred to as ``swamping'' \cite{Liu:08}.

{\it Semantic Variation:} A common aspect of many anomaly detection applications is that there can be multiple processes generating anomalies. In a cyber-security setting, there can be many different kinds of attacks and many different methods for stealing information. In cancer detection, there can be many different biological processes that result in cancerous cells. On the other hand, if there are many instances of anomalies from one generating process they may cease to appear as statistical outliers at all; such anomalies are often described as clustered anomalies. We propose \emph{Semantic Variation} as a measure of the degree to which the anomalies are generated by more than one underlying process, or, alternatively, the degree to which the anomalies are dissimilar from each other.  

When anomaly points are tightly clustered, this creates a region of high probability density, which can defeat density estimation-based methods. This phenomenon has also been called ``masking''  \cite{Liu:08}. 

{\it Relative frequency} is the fraction of the incoming data points that are anomalies of interest.  This value is the problem dimension that is most reliably reported in the literature already and has also been called ``plurality'' and ``contamination rate''. Little is done to examine the impact is has on results, however. The behavior of anomaly detection algorithms often changes with the relative frequency.  If anomalies are very rare, then methods that pretend that all training points are ``normal'' and fit a model to them may do well. If anomalies are more common, then methods that attempt to fit a model of the anomalies may do well.  In most experiments in the literature, the anomalies have a relative frequency between 0.01 and 0.1, but some go as high as 0.3 \cite{kim:08,Liu:08}. Many security applications are estimated to have relative frequencies in the range of $10^{-5}$ or $10^{-6}$. 

We believe understanding the impact of relative frequency is a fundamental issue in anomaly detection: How much can the data be contaminated by anomalies before the anomalies can no longer be reliably detected?

{\it Feature Relevance/Irrelevance:} From the application perspective, there is a natural tendency to include any feature that could conceivably be informative, but this tendency also increases the risk of including features that are irrelevant to the task.  Good knowledge of which features are germane to a task would obviate the need for an anomaly detector at all; if you know what features describe your targets, you should be able to solve your problem with either a trivial solution or an established supervised method.

It is well-established that irrelevant features can degrade the performance of supervised learning methods, and we now have many good algorithms for identifying and removing irrelevant features.  We believe that irrelevant features are an even greater problem for anomaly detection.  From the statistical perspective, each irrelevant feature increases the dimensionality of the space, and the sample size required by (naive) density estimation methods tends to scale exponentially with the dimension.  In addition, as the dimensionality of the data increases, the ``surface area'' of the volume containing the data also increases, which is a geometric way of saying that there are more ``tails'' in which the data may lie. This increases the risk that normal points will fall in the tails of the distribution.  

Because there are no labels to define the anomaly-detection task, the choice of features is the only way that the user defines the task to an anomaly detection algorithm.

\section{Benchmarking Methodology}\label{sec:method}

The goals of this study are to evaluate the impact of many factors on experimental results, including but not limited to the problem dimensions proposed in \cref{sec:require}. Because of this we acknowledge that the corpus of benchmarks generated for this study is in excess of what should be required for good experimental design. More important, making strong claims about the effects of experimental design on experimental results requires us to control benchmark creation in ways that won't always produce a good experiment; guidelines in this regard are given in \cref{sec:discuss}.

With the above in mind, the process by which we created our benchmark corpus can be briefly outlined as follows:

\begin{enumerate}
\item Select a set of existing supervised datasets (``mothersets") derived from real-world contexts.
\item Determine a ground truth label for each point; candidate normal or candidate anomaly.
\item Select points from a given motherset to construct each benchmark.
\item Attempt to vary, control and measure our four proposed problem dimensions (point difficulty, semantic variation, relative frequency and feature relevance) as benchmark creation proceeds.
\end{enumerate}

\subsection{Selecting Datasets}

To ensure objectivity in the construction of the benchmarks, we only worked with datasets from the UCI data repository \cite{uci}. We selected all UCI datasets (as of the beginning of this study) that matched the following criteria:
\begin{itemize}
\item \emph{task}: Classification (binary or multi-class) or regression. No time series.
\item \emph{instances}: At least 1000. No upper limit.
\item \emph{features}: No more than 200. No lower limit.
\item \emph{values}: Numeric only. Categorical features are ignored if
  present. No missing values, with one exception (see below).
\end{itemize}

The choice of these criteria was not guided by the performance of any anomaly detection algorithms. 

Our criteria do not cover all settings in which anomaly detection is appropriate. Instead, we focused on the common case: high-dimensional, continuous-valued, independent and identically distributed (IID) data.  Future work should explore nominal and ordinal features \cite{otey2006} as well as more structured (non-IID) settings such as time series \cite{huang2013,mei2013} and network data (e.g., \cite{bcfls-mladsgd-2014}). 

These criteria yielded a collection of 19 datasets which we refer to as the ``mothersets", since they will produce thousands of ``child" benchmark datasets. The 19 selected mothersets are the following:
\begin{itemize}
\item \emph{binary classification}: MAGIC Gamma Telescope, MiniBooNE Particle Identification, Skin Segmentation, Spambase
\item \emph{multi-class classification}: Steel Plates Faults, Gas Sensor Array Drift, Image Segmentation, Landsat Satellite, Letter Recognition, Optical Recognition of Handwritten Digits, Page Blocks, Shuttle, Waveform, Yeast
\item \emph{regression}: Abalone, Communities and Crime, Concrete Compressive Strength, Wine, Year Prediction
\end{itemize}

Communities and Crime is the one exception to our rule against missing values. In this case, there were some features for which the values where missing for the majority of points, so we simply removed these features from the data set.

In each of these mothersets, each feature was normalized to have zero mean and unit sample variance.

\subsubsection{Motherset Control}

Because this study seeks to validate many claims about experimental design, it is in our interest to perform statistical hypothesis testing where appropriate as well as provide a clear baseline for comparison in our final results. While it is well beyond the scope of this study to characterize the mothersets in ways that might explain their impact on experimental results, it is in our purview to test our claim that the selection of a motherset has any measurable impact at all. It is also in our interest to provide some evidence for the claim that benchmarks constructed from synthetic data are of a different caliber than those derived from real world data.

To this end we have created a synthetic motherset as the control group for such claims. For details of its construction, see \cref{appendix:synth}

\subsection{Defining Normal versus Anomalous Data Points}

A central goal of our methodology is that the ``normal'' and ``anomalous'' points should be produced by semantically distinct processes (Requirements 1 and 2).  To achieve this, we relabel each point in the 19 mother sets as either a candidate normal or candidate anomaly, informed by the semantics of the original set.  We employed the following methods.

\subsubsection{Binary Classification Problems}

For datasets that were already binary classification problems, the data is already partitioned into two semantically-distinct groups.  We chose one class as ``candidate normal'' (i.e., the set from which we will select the ``normal'' points) and the other as ``candidate anomaly'' (i.e., the set from which we will select the ``anomaly'' points).  The class with fewer instances is chosen  to be the candidate anomaly class. We do this because the final benchmarks will subsample the candidate anomalies so that the anomalies constitute a small fraction of all of the data points.  The larger the majority class, the easier this is to achieve. In the case that both classes are of equal size, the class with greater variance is defined to be the candidate anomaly class; we do this because the greater variance might give us additional flexibility in selecting loosely or tightly clustered anomalies at benchmark construction time.

For mother sets that are regression or multi-class problems, our approach is to transform them into binary classification problems and then treat them as just described here.  

\subsubsection{Regression Problems}

For regression datasets, the transformation into a binary problem is simple. We compute the median of the regression response and partition the data into two classes by thresholding on this value.  To the extent that low versus high values of the response correspond to different generative processes, this will create a semantic distinction between the candidate normal and candidate anomalous data points. We expect points near the median response will have high point difficulty and points near the extremes will have a low point difficulty. Benchmarks derived from regression problem sets allow for flexible (and easy) control of point difficulty.

\subsubsection{Multi-class Problems}

For multi-class datasets, we partitioned the available classes into two sets with the goal of maximizing the difficulty of telling them apart. For mother sets with many classes, it can be impractical to try every partition of the classes in the search of the most confusing binary problem, so we employ an approximation that attempts to maximize class confusion; this is the same method as used in \cite{odd2013} and is detailed again in \cref{appendix:confuse}.

\subsection{Controlling Problem Dimensions to Generate Benchmark Datasets}

The next step in our methodology is to generate benchmark datasets from each motherset by manipulating the four dimensions of point difficulty (pd), relative frequency (rf), clusteredness (cl), and feature irrelevance (fi). In this section, we define  a continuous measure for each of these dimensions. We then define a set of levels for each measure used during construction.  Each benchmark dataset corresponds to choosing one level of each factor, (or its control setting).  For each mother set, we iterate over each combination of problem dimension levels and construct 5 benchmark datasets having those settings. As we will describe below, limitations of some mother datasets mean that we cannot always achieve 5 datasets for every desired setting.

In total, from the 19 mother sets listed earlier, this methodology produced 25,685 benchmark datasets which form the corpus used for this study.

Benchmarks are constructed by sampling points from the motherset one at a time while observing feasibility constraints on selecting a point that do not violate the desired problem dimension settings. How these constraints are managed for each dimension are detailed below. For any particular problem dimension, creating a control group for that dimension is simply a matter of not enforcing the corresponding constraint.

Additionally, benchmarks observe some size constraints as well. To ensure variability among the candidate normals used in each benchmark, we set a hard cap of 90\% of available candidate normals for any benchmark. We also set a global cap of 6,000 points used in a single benchmark, this value chosen in part to accommodate algorithms that do not scale well to large data sets.

\subsubsection{Point Difficulty}

We propose to define point difficulty (pd) based on an oracle that knows the true generating processes underlying the ``normal'' and ``anomalous'' points.  Using this knowledge, the oracle can estimate the probability $P(y=\mbox{normal}|x)$ that a data point $x$ was generated by the ``normal'' distribution or $P(y=\mbox{anomaly}|x)$ that a data point $x$ was generated by the ``anomalous'' distribution.  We consider the point difficulty of any point to be the estimated probability that it belongs to the other class. The point difficulty level of an entire benchmark is summarized as the mean of the point difficulty of all the points in the benchmark.

To arrive at these probability estimates, we applied Kernel Logistic Regression (KLR; \cite{klr1,klr2,klr3}) as our oracle. Specifically, we implemented the algorithm described by Keerthi, et al., \cite{klr3} with parameters chosen via 5-fold cross-validation. On very large mothersets the KLR models underwent a further approximation by subsampling the points used to define our kernel space.

Our measure of point difficulty is a measure that is sensitive the distance from the decision boundary between the normal and anomalous distributions, rather than some measure that would be sensitive to the euclidean distances between the anomalous and normal points.

We binned this measure into five discrete levels:
\begin{itemize}
\item \emph{pd-0}: control group; (difficulty score $\in(0,1))$
\item \emph{pd-1}: difficulty score $\in(0,0.1\overline{6})$
\item \emph{pd-2}: difficulty score $\in[0.1\overline{6},0.\overline{3})$
\item \emph{pd-3}: difficulty score $\in[0.\overline{3},0.5)$
\item \emph{pd-4}: difficulty score $\in[0.5,1)$
\end{itemize}

When creating benchmarks at a particular point difficulty setting, we enforce this level by not selecting points that would move our mean point difficulty outside of this range.

Although we doubt that benchmarks limited to pd-4 will resemble any real application domain (as these benchmarks would almost entirely consist of points that confused even a supervised oracle) it is in our interest to test the extremes of each setting in our study.

\subsubsection{Relative Frequency}

This setting is very easily controlled; if we desire that a benchmark is 0.01 contaminated with anomalies, then we simply ensure that the benchmark draws 0.99 of its points from the candidate normals and 0.01 of its points from the candidate anomalies. We defined the following six relative frequency (rf) levels:

\begin{itemize}
\item \emph{rf-0}: control group; (contamination rate is not considered).
\item \emph{rf-1}: 0.001 of the benchmark is drawn from the candidate anomalies.
\item \emph{rf-2}: 0.005 of the benchmark is drawn from the candidate anomalies.
\item \emph{rf-3}: 0.01 of the benchmark is drawn from the candidate anomalies.
\item \emph{rf-4}: 0.05 of the benchmark is drawn from the candidate anomalies.
\item \emph{rf-5}: 0.1 of the benchmark is drawn from the candidate anomalies.
\end{itemize}

Points for benchmarks are selected one at a time regardless of their ground truth label. The relative frequency setting is used to determine a limit on the number of candidate anomalies or candidate normals that can be sampled; once one class is ``used up" selecting points from that ground truth setting is no longer feasible. During construction it is possible that we will run out of feasible points before both limits are reached. When this happens, previously selected points are removed from the benchmark (while still observing other feasibility settings) until the appropriate relative frequency is achieved.

It is worth noting the the control group for this setting is to not consider the relative frequency at all which often (though not always) results in benchmarks that have an unrealistically large portion of ``anomalies''. However, we cannot make strong claims about the impact of manipulating relative frequency if we don't also demonstrate the impact of \emph{not} manipulating it.

\subsubsection{Semantic Variation and Clusteredness}

We propose to define semantic variation among anomalies with the following measure. The {\it normalized clusteredness} (nc) of this set of points is defined as 
\[\log\left(\frac{\hat{\sigma}^2_n}{\hat{\sigma}^2_a}\right)\]
where $\hat{\sigma}^2_n$ is the sample variance of the selected normal points and $\hat{\sigma}^2_a$ is the sample variance of the selected anomaly points.  When normalized clusteredness is less than 0, the anomaly points exhibit greater semantic variation than the normal points. When normalized clusteredness is greater than 0, the anomaly points are more tightly packed than the normal points (on average).

While we would prefer to have greater control over this measure, maintaining the feasibility of the other problem dimensions is sometimes at odds with trying to maximize or minimize this quantity, so we defined only the three normalized clusteredness levels at construction time:

\begin{itemize}
\item \emph{nc-0}: control group; (clusteredness not considered).
\item \emph{nc-1}: nc $< 0$; (scattered anomalies).
\item \emph{nc-2}: nc $> 0$; (clustered anomalies).
\end{itemize}

When selecting a point, points that would change the current measure of normalized clusteredness to the wrong side of 0 are not considered feasible. However we would like greater variety among the final clusteredness measure, so the selection of points gives greater weight to points that would push this measure further from zero.

\subsubsection{Feature Irrelevance}

In the terms discussed earlier, it is unknown to us how relevant the features of the motherset are to the task of detecting targeted outliers, so we assume that the original data offers the most compact set of features for this purpose. There is no feasibility constraint for this dimension. Instead we control our measure of feature irrelevance by adding irrelevant features to an otherwise finished benchmark until a desired level of feature irrelevance is introduced.

We quantify the amount of feature irrelevance as follows. First, we measure the average distance between all pairs of points in the benchmark dataset prior to adding irrelevant features. Then we add irrelevant features until the average pairwise distance has increased by a desired ratio.  We define four levels of feature irrelevance (fi):
\begin{itemize}
\item \emph{fi-0}: control group; average distance ratio of 1.0 (no added irrelevant features).
\item \emph{fi-1}: Average distance ratio of 1.2
\item \emph{fi-2}: Average distance ratio of 1.5
\item \emph{fi-3}: Average distance ratio of 2.0
\end{itemize}

To create a new irrelevant feature, we choose a feature from the original motherset uniformly at random. Then for each data point in the benchmark dataset, we choose a value for this feature by sampling uniformly (with replacement) from the values of the original data points.  The result is that this new feature has the same marginal distribution as some original feature, but its values carry no information about the anomaly status of the data points.  This preserves the idiosyncrasies of the real data while allowing us to introduce noise.

To simplify the process of determining how many irrelevant features are needed, we compute an estimate of how many extra features will achieve a desired average distance ratio. Note that the expected distance between two vectors ($\alpha$) whose coordinates are drawn at random (e.g., from the unit interval or from a standard normal Gaussian) grows in proportion to $\sqrt{d}$, where $d$ is the dimensionality of the data. Hence, if a dataset already has $d$ dimensions and we want to estimate $d'$, the number of dimensions needed to increase the average pairwise distance by a factor of $\alpha$, then we need $$\hat{d'} = (\alpha \sqrt{d})^2$$
dimensions, where $\alpha\in\{1.0,1.2,1.5,2.0\}$ for this study.

\newpage

\subsection{Available Software}

For greater clarity on the above process, the software used to produce our corpus of benchmarks is available at

\ 

\url{http://ir.library.oregonstate.edu/xmlui/handle/1957/59114}

\ 

Also available is the corpus of benchmarks itself.

Our final corpus of benchmarks provides what we believe to be a rich environment to analyze the anomaly detection problem.

\section{Anomaly Detection Algorithms}\label{sec:algo}

As part of our study we applied eight anomaly detection algorithms to the entire corpus of benchmarks. This selection of algorithms is not exhaustive but we believe it is a good representation of the field and seeks to cover both classic and state-of-the-art algorithms while also trying to cover several different solution approaches. This section only summarizes each algorithm and why it was selected. For specific implementation and paramaterization details refer to \cref{appendix:parameters}.

\subsection{Density-Based Approaches}

A straightforward way to generate outlier scores is to estimate some probability distribution fit to the data points; the less likely the point, the more likely it is an anomaly. Such approaches can be strongest when the outliers-as-target assumption is valid; that is, when the set of features in use are more relevant to the task.

\subsubsection{Robust Kernel Density Estimation (rkde)}

One approach to fitting a flexible, non-parametric probability density model is to fit distributions with simple parameters to each point in the data and then build an additive model that combines them. Such methods are known to be sensitive to outliers and so more complicated methods from robust statistics have been introduced. Kim, et al. \cite{kim:08} describe the strategies for optimizing kernel density estimates that we applied in this study.

The outlier scores produced are the negative log-likelihood of each point according to the final model.

\subsubsection{Ensemble Gaussian Mixture Model (egmm)}

Another approach to density estimation is to fit a Gaussian mixture model (GMM) using the EM algorithm. A single GMM is not very robust, and like $k$-means clustering it requires that we select a value of $k$ Gaussian mixture components. To improve robustness, we computed an ensemble of GMMs for many values of $k$, discarded models that did not fit the data well, and then combined the predicted densities of the remaining models. As above, the outlier scores produced are based on the negative log-likelihood of each point according to the final models.

\subsection{Model-Based Approaches}

Another anomaly detection strategy is to assume that the vast majority of points are normal and thus constructing a model of the data should produce a descriptive decision boundary.

\subsubsection{One-Class SVM (ocsvm)}

The One-Class SVM algorithm (Scholkopf et al. \cite{Scholkopf:99}) uses a Support-Vector Machine to search for a kernel-space decision boundary that separates fraction $1-\delta$ of the data from the kernel-space origin. The outlier scores produced by this algorithm are determined by the residual after each point is projected onto the decision surface. Points outside the decision boundary will have positive residuals, where interior points will have negative residuals.

\subsubsection{Support Vector Data Description (svdd)}

As proposed by Tax and Duin \cite{Tax:2004} and similar in concept to One-Class SVM, Support Vector Data Description finds the smallest hypersphere in kernel space that encloses $1-\delta$ of the data. As above, the outlier scores produced by this algorithm are determined by the residual after each point is projected onto the decision surface.

\subsection{Nearest Neighbors-Based Approaches}

Often called distance-based approaches, we offer the alternative name because KNN Angle-Based Outlier Detection does not actually use distance in the computation of its anomaly score. Like LOF, however, it assumes that information about the local character of the space surrounding a point can be found by examining its neighbors, which in some ways lets euclidian distance function as a proxy for density estimation.

\subsubsection{Local Outlier Factor (lof)}

The well-known Local Outlier Factor algorithm (Breunig, et al. \cite{Breunig:2000}) computes the outlier score of a point $x_i$ by computing its average distance to its $k$ nearest neighbors. It normalizes this distance by computing the average distance of each of those neighbors to {\it their} $k$ nearest neighbors.  So, roughly speaking, a point is believed to be more anomalous if it is significantly farther from its neighbors than they are from each other.

\subsubsection{KNN Angle-based Outlier Detection (abod)}

Angle-Based Outlier Detection as proposed by Kriegel, et al in \cite{abod} in its full form is an algorithm of cubic complexity as follows. For each point $x_i$, consider all pairs of other points $(x_j,x_k)\in X,i\ne j\ne k$ and compute the angle between them relative to to point $x_i$. The sample variance of these angles determines the outlier score, with \emph{lower} variances indicating anomalous points. Because of the run-time complexity, two simple approximations were suggested by the authors. The first is to sub sample the data and use this as the reference set for computing angles. The other is to only consider the angles among the $k$ nearest neighbors to $x_i$. In initial experiments we found the latter to outperform the former and so that is the strategy employed in this study.

\subsection{Projection-Based Approaches}

Typically Isolation Forest has stood alone among well-established ``Isolation-Based'' techniques but we instead group it with the newer LODA algorithm because they both rely on using information from random projections of the data, albeit in much different ways.

\subsubsection{Isolation Forest (iforest)}

The Isolation Forest algorithm (Liu, et al. \cite{Liu:08}) ranks data points as anomalous if they are easily isolated by random axis-parallel splits. An isolation forest is a set of isolation trees where each isolation tree is an extremely random decision tree where each decision point is determined by choosing a feature uniformly at random and then choosing a splitting threshold uniformly at random between the minimum and maximum observed values of that feature. The intuition is that outliers are points that are easily isolated at random from the remaining data, so points with small average isolation depth are likely to be anomalies and so are assigned high anomaly scores.

\subsubsection{Lightweight Online Detector Of Anomalies (loda)}

As proposed by Pevny in \cite{loda}, LODA generates several ``weak'' anomaly detectors by producing many random projections of the data and then computing a density estimation histogram for each projection. The outlier scores produced are the mean negative log-likelihood according to each histogram for each point.

\section{Evaluation Metrics and Hypothesis Tests}\label{sec:hypo}

With eight algorithms and 25,685 benchmark datasets we have results from 205,480 micro-experiments which need to be evaluated. The two most common evaluation metrics in the literature are area under the ROC curve (AUC) and Average Precision (AP). Area under the Precision-Recall curve is identical to AP if the curve does not use the trapezoidal interpretation between true positives.

Precision-at-$k$ or Recall-at-$k$ metrics are also common but run the risk of being application specific (the appropriate selection of $k$ is determined by the context of the application) and we feel not appropriate for a broad meta-analysis such as this, but we note that such metrics might be the more useful in a real-world setting.

Because we tested many algorithms under many conditions (some of them unfairly difficult) there will exist many results that seem poor. We want some certainty that observed differences in results are not just random noise. For both AUC and AP, we performed an hypothesis test on each micro-experiment output, with the null hypothesis in each case being that the algorithm's output was random and the alternative being that it was not. 

Specifically, one can treat the AUC and AP of a random ranking of points as random variables each with parameters $n_{\text{anom}}$ (the number of anomalies in the benchmark) and $n_{\text{norm}}$ (the number of normals) and then compute the quantiles of interest for each of these distributions. Conducting a test with significance $\alpha$ is a matter of computing the $(1-\alpha)$-quantile of the appropriate random variable. Refer to \cref{appendix:random} for an overview of how AUC and AP can be treated as random variables.

A more intuitive view of this process: if algorithm $a$ achieves an AUC 0f 0.75 on benchmark $b$ and we reject the null hypothesis with $\alpha=0.001$, that would mean that with probability at least 0.999 a random ranking would achieve an AUC worse than 0.75 \emph{on benchmark $b$}.

We performed such hypothesis tests for both AUC and AP and for $\alpha\in(0.05,0.01,0.001)$ for each micro-experiment result. Failure in this context refers to failing to reject the null hypothesis. To summarize our tests, we first define the notion of \textbf{benchmark failure} as a benchmark instance for which \emph{all} algorithms failed. While benchmark failure is dependent on the algorithms used in this study, we still believe it is a good indication that we should not use the benchmark as evidence for later conclusions.

\subsection{Results of Hypothesis Tests by Benchmark Construction Factors}

\cref{tbl:hypo} summarizes the global benchmark failure rates for AUC and AP. The ``Either'' column indicates benchmarks for which all algorithms failed under at least one of the two metrics.
\begin{table}
\resizebox{\textwidth}{!}{\begin{minipage}{0.5\textwidth}
\tbl{Benchmark Failure Rate by Metric and Significance Level\label{tbl:hypo}}{
\begin{tabular}{|r||c|c|c|}
\hline
&\textbf{AUC}&\textbf{AP}&\textbf{Either}\\
\hline\hline
$\alpha=0.05$&0.2418&0.2815&0.3337\\
$\alpha=0.01$&0.3282&0.4162&0.4609\\
$\alpha=0.001$&0.4108&0.5087&0.5430\\
\hline
\end{tabular}}
\end{minipage}}
\end{table}

The appropriate significance level for this study is debatable. Smaller $\alpha$ trades away potential evidence (by eliminating benchmarks) for greater confidence that the results from the benchmarks under consideration are relevant. We choose to apply the more stringent threshold of $\alpha=0.001$; even though the failure rate at this level is rather high, it still leaves many benchmarks across all factors of interest.

\cref{tbl:msetfail,tbl:origfail,tbl:arfail,tbl:pdfail,tbl:clfail,tbl:irfail} show the benchmark failure rates across our various benchmark construction criterion. Boldface values indicate a failure rate greater than the global average for that metric, suggesting that factor may be unsuitable for benchmark construction, or at least should be used with caution.

\begin{table}
\resizebox{\textwidth}{!}{\begin{minipage}{0.5\textwidth}
\tbl{Benchmark Failure Rate by Metric and Motherset ($\alpha=0.001$)\label{tbl:msetfail}}{
\begin{tabular}{|r||c|c|c|}
\hline
&\textbf{AUC}&\textbf{AP}&\textbf{Either}\\
\hline\hline
synthetic&0.2417&0.2553&0.2765\\
abalone&0.3838&0.4533&0.4841\\
comm.and.crime&\textbf{0.4924}&0.5053&\textbf{0.5727}\\
concrete&\textbf{0.5121}&\textbf{0.5908}&\textbf{0.6377}\\
fault&\textbf{0.4178}&\textbf{0.6105}&\textbf{0.6283}\\
gas&0.2767&\textbf{0.5178}&0.5189\\
imgseg&0.2792&0.2569&0.3667\\
landsat&\textbf{0.5286}&\textbf{0.5617}&\textbf{0.6378}\\
letter.rec&\textbf{0.6276}&\textbf{0.7273}&\textbf{0.7360}\\
magic.gamma&0.3300&0.3322&0.3917\\
opt.digits&\textbf{0.5567}&\textbf{0.7115}&\textbf{0.7364}\\
pageb&0.0468&0.1894&0.1915\\
particle&0.2533&0.3678&0.4306\\
shuttle&0.0944&0.2711&0.2722\\
skin&0.0853&0.3673&0.3693\\
spambase&0.3622&0.4844&0.5178\\
wave&\textbf{0.4954}&\textbf{0.6000}&\textbf{0.6278}\\
wine&\textbf{0.4860}&\textbf{0.6355}&\textbf{0.6554}\\
yearp&\textbf{0.6822}&\textbf{0.7239}&\textbf{0.7572}\\
yeast&\textbf{0.9733}&\textbf{0.9789}&\textbf{0.9900}\\
\hline
\end{tabular}}
\end{minipage}}
\end{table}

\begin{table}
\resizebox{\textwidth}{!}{\begin{minipage}{0.5\textwidth}
\tbl{Benchmark Failure Rate by Metric and Motherset Origin ($\alpha=0.001$)\label{tbl:origfail}}{
\begin{tabular}{|r||c|c|c|}
\hline
&\textbf{AUC}&\textbf{AP}&\textbf{Either}\\
\hline\hline
binary&0.2490&0.3530&0.3914\\
multiclass&\textbf{0.4488}&\textbf{0.5627}&\textbf{0.5913}\\
regression&\textbf{0.5159}&\textbf{0.5830}&\textbf{0.6219}\\
\hline
\end{tabular}}
\end{minipage}}
\end{table}

\begin{table}
\resizebox{\textwidth}{!}{\begin{minipage}{0.5\textwidth}
\tbl{Benchmark Failure Rate by Metric and Relative Frequency Level ($\alpha=0.001$)\label{tbl:arfail}}{
\begin{tabular}{|r||c|c|c|}
\hline
&\textbf{AUC}&\textbf{AP}&\textbf{Either}\\
\hline\hline
rf-0&0.3432&0.4212&0.4324\\
rf-1&\textbf{0.7286}&\textbf{0.8464}&\textbf{0.9081}\\
rf-2&\textbf{0.5432}&\textbf{0.6889}&\textbf{0.7492}\\
rf-3&\textbf{0.4395}&\textbf{0.5786}&\textbf{0.6395}\\
rf-4&0.2350&0.3076&0.3234\\
rf-5&0.2533&0.3031&0.3139\\
\hline
\end{tabular}}
\end{minipage}}
\end{table}

\begin{table}
\resizebox{\textwidth}{!}{\begin{minipage}{0.5\textwidth}
\tbl{Benchmark Failure Rate by Metric and Point Difficulty Level ($\alpha=0.001$)\label{tbl:pdfail}}{
\begin{tabular}{|r||c|c|c|}
\hline
&\textbf{AUC}&\textbf{AP}&\textbf{Either}\\
\hline\hline
pd-0&0.2887&0.3951&0.4328\\
pd-1&0.2803&0.3988&0.4268\\
pd-2&\textbf{0.4252}&\textbf{0.5257}&\textbf{0.5576}\\
pd-3&\textbf{0.5662}&\textbf{0.6481}&\textbf{0.6858}\\
pd-4&\textbf{0.7540}&\textbf{0.8041}&\textbf{0.8437}\\
\hline
\end{tabular}}
\end{minipage}}
\end{table}

\subsection{Results of Hypothesis Tests by Algorithm}

For a simple comparison of algorithms, \cref{tbl:algofail} shows the failure rate of each algorithm across the entire corpus, with the lowest failure rate in each metric boldfaced.

\begin{table}
\resizebox{\textwidth}{!}{\begin{minipage}{0.5\textwidth}
\tbl{Benchmark Failure Rate by Metric and Clustering Strategy ($\alpha=0.001$)\label{tbl:clfail}}{
\begin{tabular}{|r||c|c|c|}
\hline
&\textbf{AUC}&\textbf{AP}&\textbf{Either}\\
\hline\hline
none&\textbf{0.4189}&\textbf{0.5201}&\textbf{0.5479}\\
cluster&\textbf{0.4431}&\textbf{0.5891}&\textbf{0.6036}\\
scatter&0.3695&0.4146&0.4759\\
\hline
\end{tabular}}
\end{minipage}}
\end{table}

\begin{table}
\resizebox{\textwidth}{!}{\begin{minipage}{0.5\textwidth}
\tbl{Benchmark Failure Rate by Metric and Feature Irrelevance Level ($\alpha=0.001$)\label{tbl:irfail}}{
\begin{tabular}{|r||c|c|c|}
\hline
&\textbf{AUC}&\textbf{AP}&\textbf{Either}\\
\hline\hline
ir-0&0.3065&0.4008&0.4327\\
ir-1&0.3785&0.4760&0.5098\\
ir-2&\textbf{0.4356}&\textbf{0.5362}&\textbf{0.5743}\\
ir-3&\textbf{0.5224}&\textbf{0.6215}&\textbf{0.6549}\\
\hline
\end{tabular}}
\end{minipage}}
\end{table}

\begin{table}
\resizebox{\textwidth}{!}{\begin{minipage}{0.5\textwidth}
\tbl{Algorithm Failure Rate by Metric ($\alpha=0.001$)\label{tbl:algofail}}{
\begin{tabular}{|r||c|c|c|}
\hline
&\textbf{AUC}&\textbf{AP}&\textbf{Either}\\
\hline\hline
abod&0.5898&0.6784&0.7000\\
iforest&\textbf{0.5520}&\textbf{0.6514}&\textbf{0.6741}\\
loda&0.6187&0.6955&0.7194\\
lof&0.6016&0.7071&0.7331\\
rkde&0.6122&0.7030&0.7194\\
ocsvm&0.7218&0.7342&0.7960\\
svdd&0.8482&0.8868&0.9080\\
egmm&0.6188&0.7146&0.7303\\
\hline
\end{tabular}}
\end{minipage}}
\end{table}

We defer drawing conclusions about these failure rates until \cref{sec:discuss} but will point out a few details now. One motherset in particular, \textbf{yeast}, has a failure rate approaching 100\%, a strong indication that it is a very poor choice for benchmark construction. Also, failure rates across relative frequency levels need to be interpreted more carefully because the threshold for rejecting the null hypothesis is itself impacted by the relative frequency.

\section{Results Summary}\label{sec:basic}

We now examine factors contributing to micro-experiment results more closely. For each metric, we eliminate all results from failed benchmarks (This \emph{does} mean that reported results for AUC and AP will be drawing from somewhat different pools of benchmarks). We retain individual algorithm failures; as long as the benchmark has at least one success we have evidence that a failing algorithm could have distinguished itself from random behavior and so its failure to do so is relevant.

\subsection{Transformation of Metrics}

Later in this section we will to analyze our micro-experiment results with various linear models, and summarizing the means of metrics like AUC and AP which are constrained to range $[0,1]$ can be problematic, especially in the case of AP which does not have a constant expectation. For the remainder of this paper we transform both of these metrics so that they extend to the range of all real numbers in the following ways:

\begin{itemize}
\item \textbf{AUC}-- We use the logit transform:
$$\text{logit}(AUC)=\log\left(\frac{AUC}{1-AUC}\right)$$
\item \textbf{AP}-- Because AP does not have a constant expectation, one way to normalize AP is to compute the \textbf{lift} which is the ratio of AP to its expectation. It is commonly assumed that this expectation is equivalent to the anomaly rate, but \cite{exactAP} shows that while this can be a good approximation it is not exactly correct. More important it can be a bad approximation when relative frequency is low, which is the case for most of our benchmarks. We compute lift using the exact expectation. To map this ratio to all real numbers a sensible further transformation is to take the log of this lift:
$$\log(LIFT)=\log\left(\frac{AP}{E[AP]}\right)$$
\end{itemize}

Note that both of these transformed metrics have an expectation of zero.

\subsection{Impact of Benchmark Construction Criterion}

\begin{sidewaysfigure}
\caption{}\label{fig:mset}
\includegraphics[width=\textwidth]{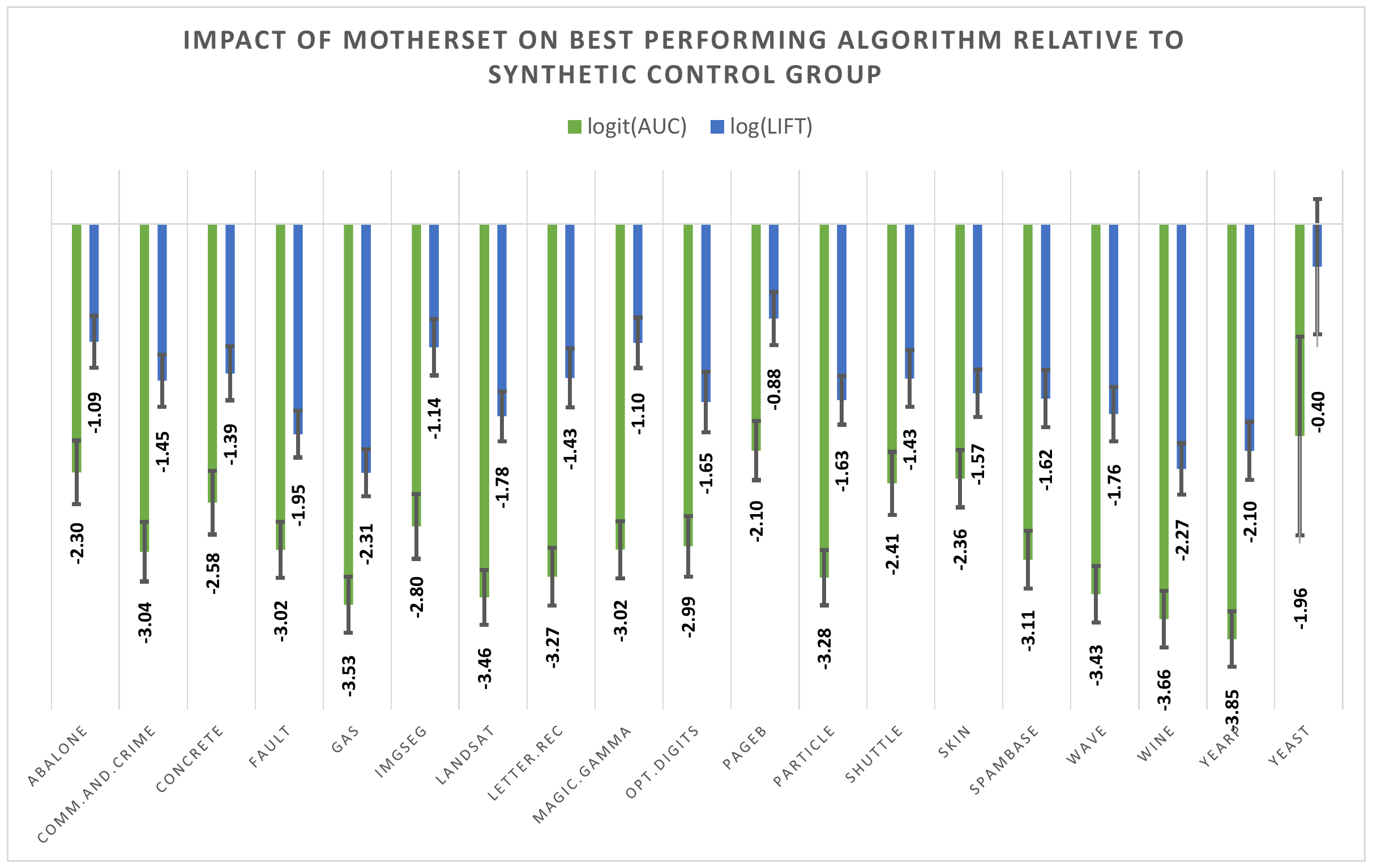}
\end{sidewaysfigure}

\begin{figure}
\caption{}\label{fig:ar}
\includegraphics[width=\textwidth]{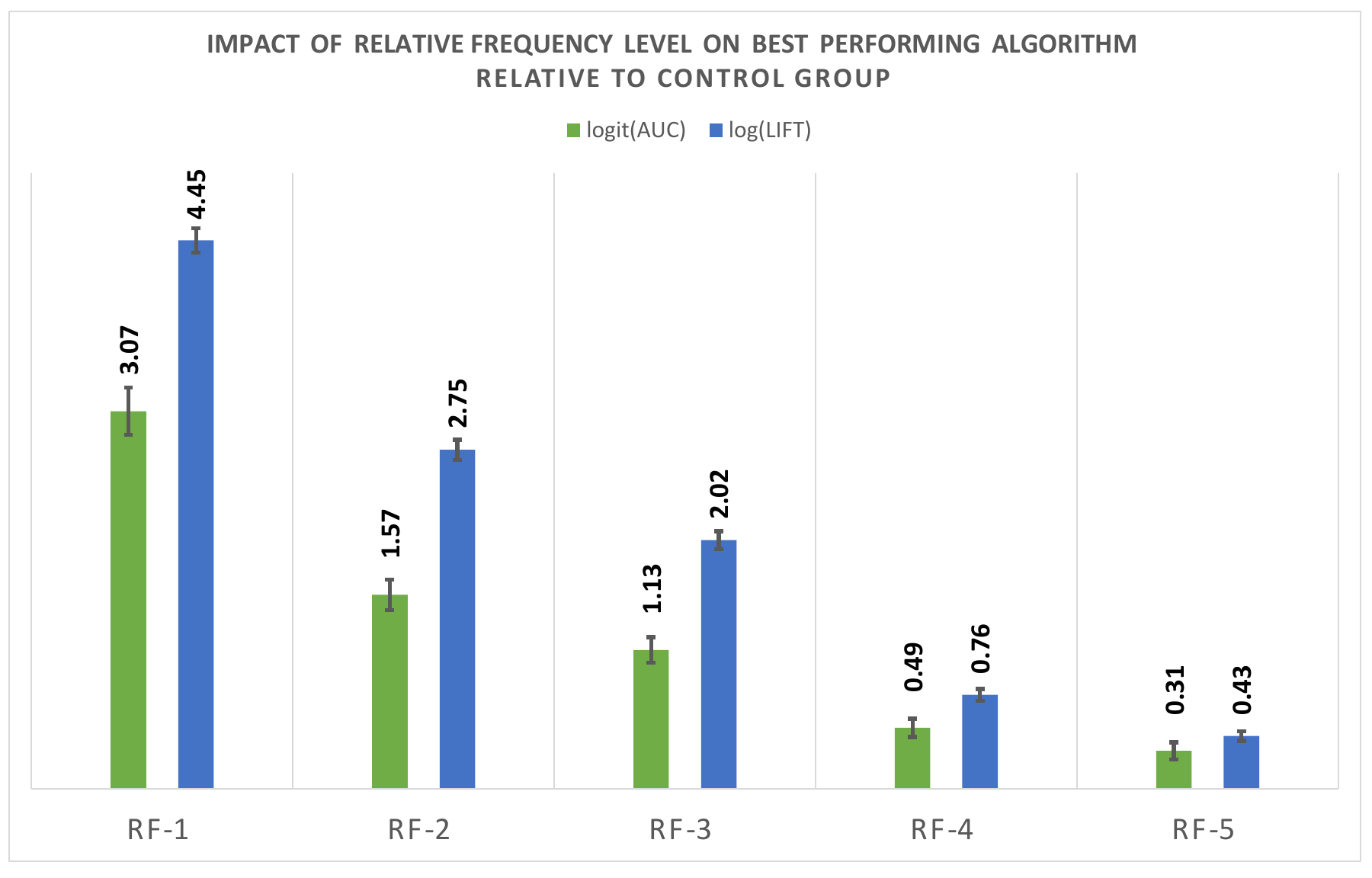}
\caption{}\label{fig:pd}
\includegraphics[width=\textwidth]{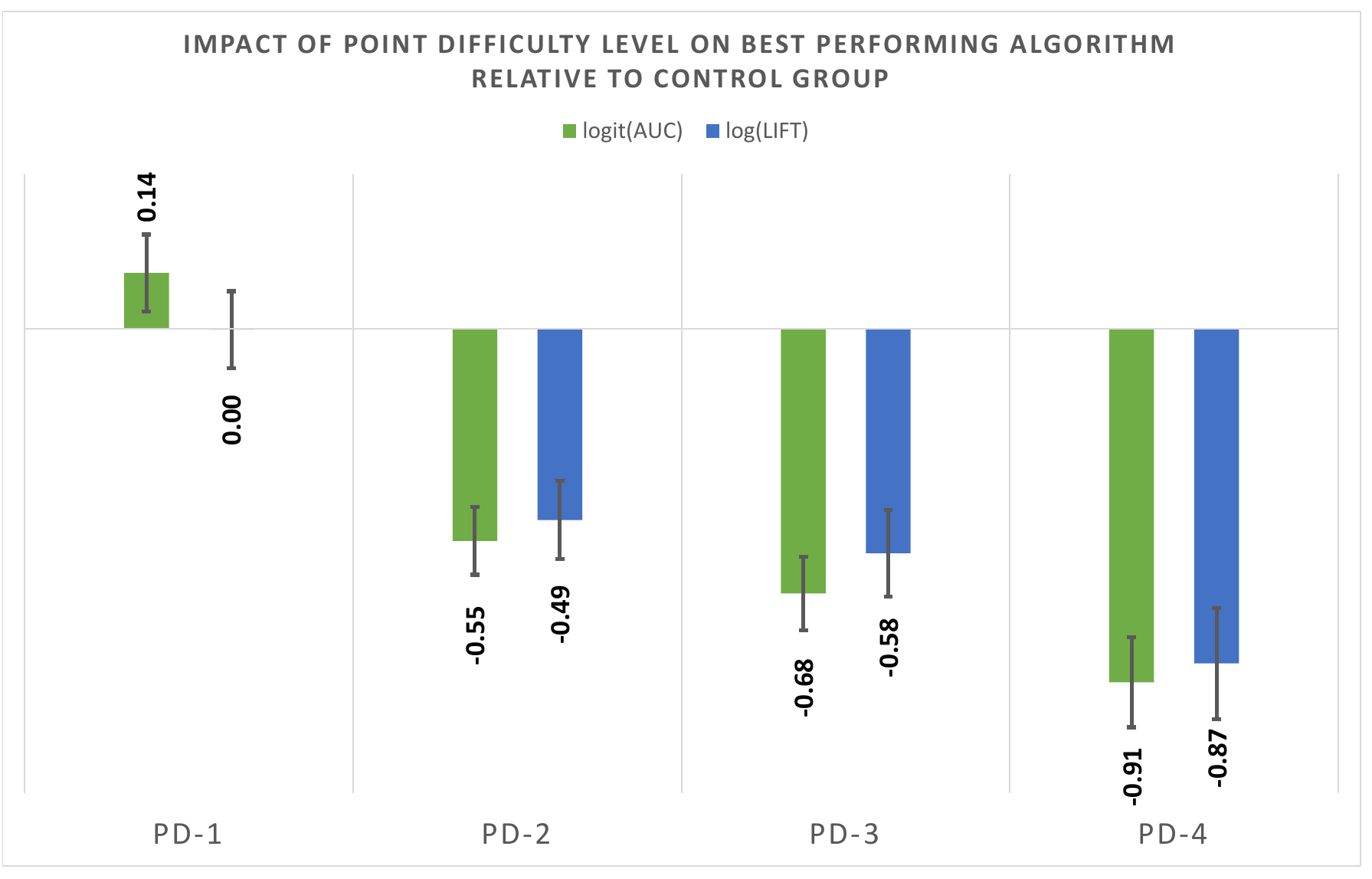}
\end{figure}

\begin{figure}
\caption{}\label{fig:cl}
\includegraphics[width=\textwidth]{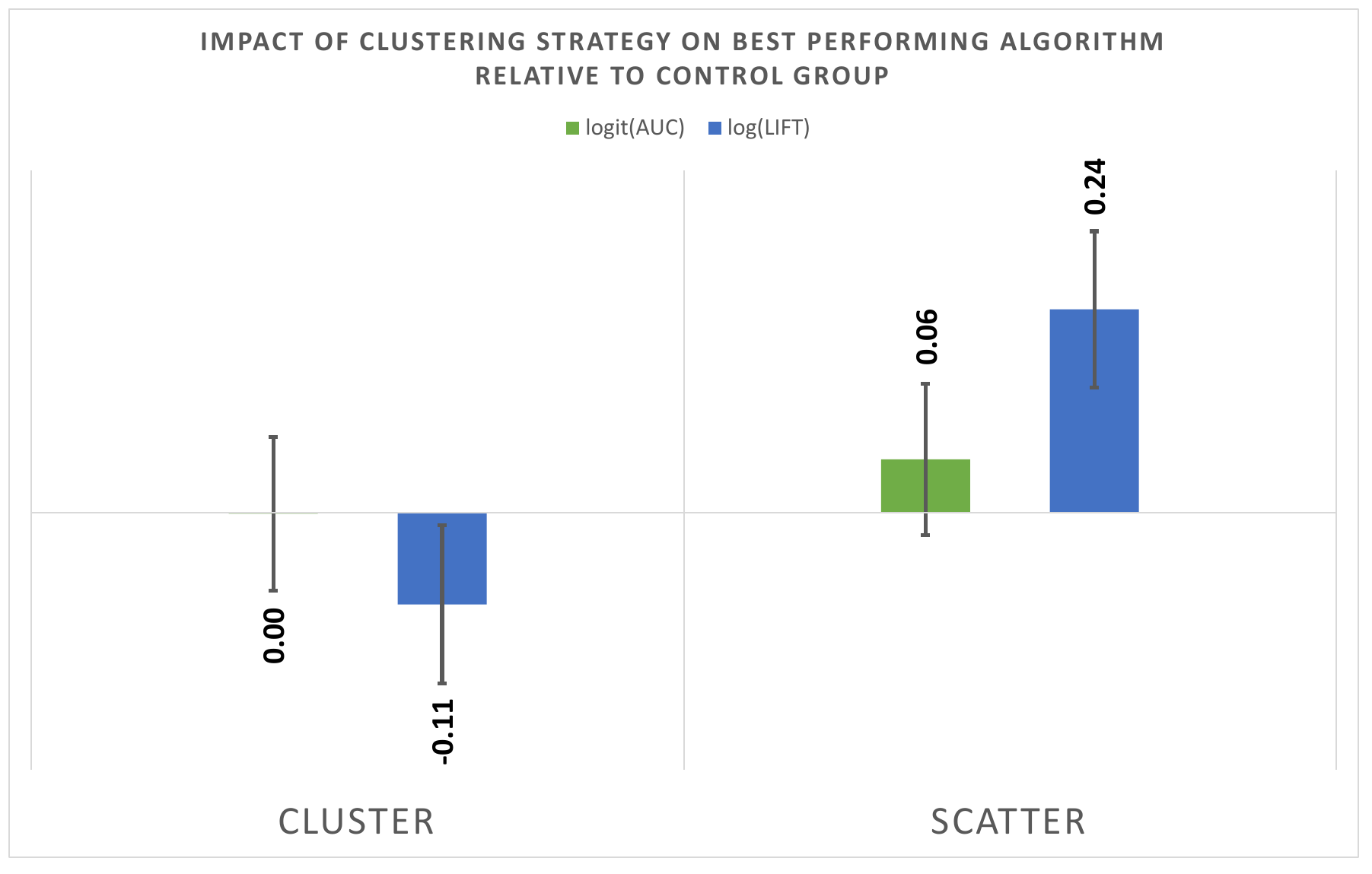}
\caption{}\label{fig:ir}
\includegraphics[width=\textwidth]{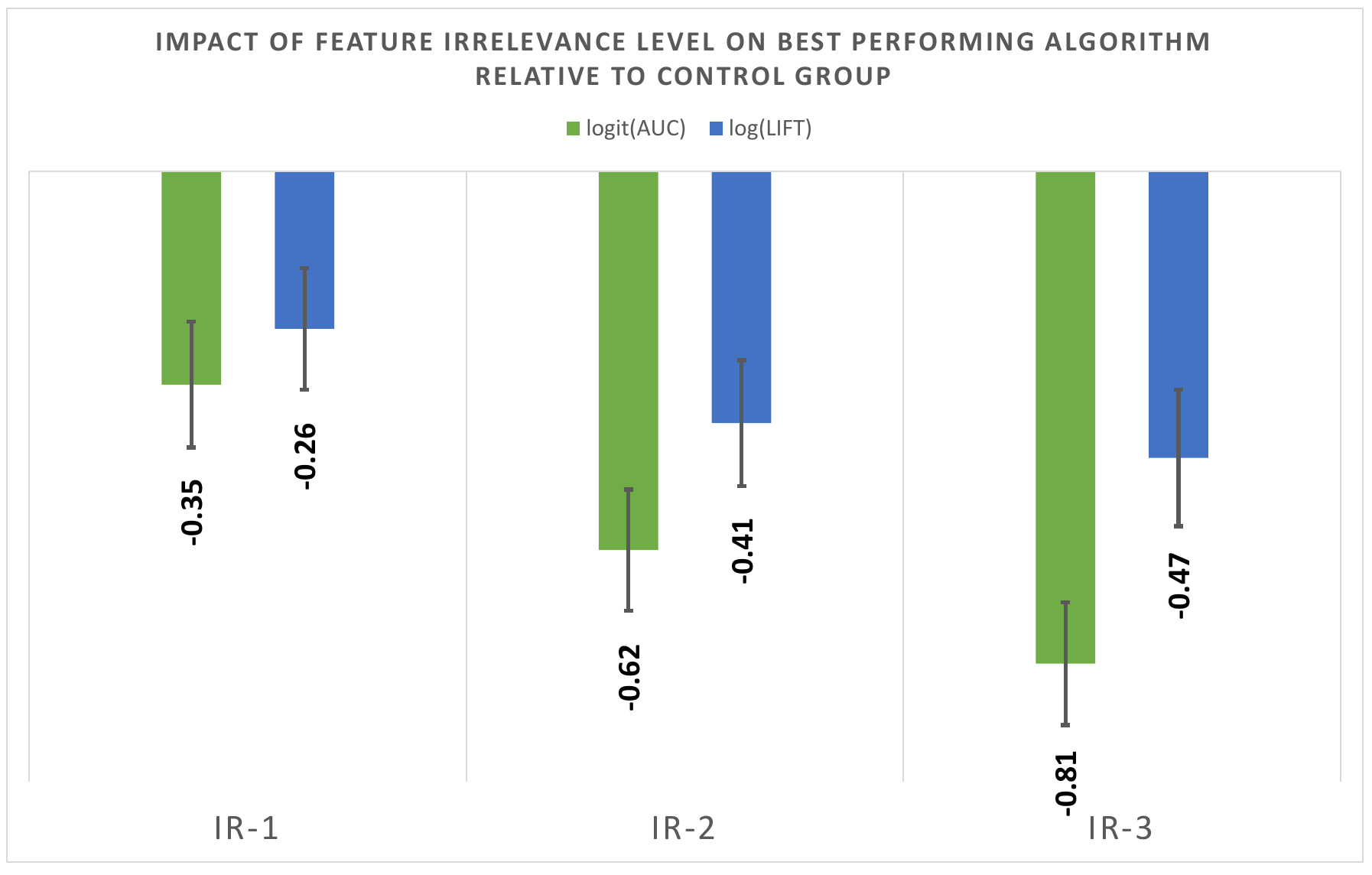}
\end{figure}

To demonstrate the impact of our benchmark construction methodology on micro-experiment results we first want to examine results in a way that is agnostic to choice of algorithm. Because we are only examining results from benchmarks where at least one algorithm produced statistically significant output, we know that if we only consider the best result from each benchmark we are always choosing a result that is better than random with high confidence. As each benchmark construction factor has a well defined control group, we compute the mean difference in performance-of-best-algorithm between each level and the control group, and then place a $0.999$ confidence interval around this difference. The results are displayed in \cref{fig:mset,fig:ar,fig:pd,fig:cl,fig:ir}. Observe that the metrics logit(AUC) and log(LIFT) are not meant to be compared to each other, but are shown side by side for a more compact presentation.

Of note in \cref{fig:mset} is that all mothersets had a negative impact relative to the synthetic control group. Of course, the synthetic control group is of our own design and such a synthetic control could be produced in several ways, some of which might generate more challenging benchmarks, but the general trend should indicate that using real data is notably more challenging than synthetic toy problems. However, the \textbf{yeast} motherset does not statistically distinguish itself from the control group, but this can be mostly disregarded as nearly all \textbf{yeast} benchmarks were discarded due to the motherset's very large failure rate; we have already more or less disqualified the \textbf{yeast} set from consideration anyway.

\cref{fig:ar,fig:pd,fig:ir} all show a trend of algorithm performance decreasing as the problem dimension's relevant quantity is increased. Note in \cref{fig:pd}, however, that the easiest point difficulty setting is not distinguishable from the control setting.

\cref{fig:cl} does suggest that scattered anomalies are easier than clustered anomalies, but neither setting is distinguishable from the control setting. As a discrete factor, the choice of clustering algorithm may not be adequate to measure the impact of clusteredness on algorithm performance, but we will show later that clusteredness in general does have a statistically significant impact.

We state now that \cref{fig:mset,fig:ar,fig:pd,fig:cl,fig:ir} are evidence of the impact of our methodology on overall benchmark difficulty. The greater impact of our problem dimensions will be examined later in this section.

\subsection{Simple Comparison of Algorithms}

Again, for a simple comparison of algorithms, \cref{tbl:meanalgo} shows the mean performance of each algorithm by each of our transformed metrics. The best result in each metric is bolded.

For some intuition on these scores: \textbf{iforest's} logit(AUC) of 1.0893 is analogous to an AUC of 0.7482 and \textbf{svdd's} score of 0.1538 is analogous to an AUC of 0.5384. \textbf{iforest's} log(LIFT) of 1.0918 is analogous to a lift of 2.9796 (roughly 3 times better than expectation) and \textbf{svdd's} score of 0.2806 is analogous to a lift of 1.3239 (roughly 1.3 times better than expectation). 

\begin{table}
\resizebox{\textwidth}{!}{\begin{minipage}{0.5\textwidth}
\tbl{Mean Performance by Algorithm and Metric\label{tbl:meanalgo}}{
\begin{tabular}{|r||c|c|}
\hline
&\textbf{Mean logit$(AUC)$}&\textbf{Mean $\log(LIFT)$}\\
\hline\hline
abod&0.9517&0.9009\\
egmm&0.9678&0.9081\\
iforest&\textbf{1.0893}&\textbf{1.0918}\\
loda&0.8604&0.9156\\
lof&0.9632&0.9329\\
ocsvm&0.5650&0.8608\\
rkde&0.9554&0.9132\\
svdd&0.1538&0.2806\\
\hline
\end{tabular}}
\end{minipage}}
\end{table}

\subsection{Analyzing Results with Linear Regression Models}

Our discrete benchmark construction factors can only provide a granular view of overall results and we should like to include results from all algorithms not just the best algorithm at each benchmark. To this end we constructed several fixed and mixed effect models to explain logit(AUC) and log(LIFT) across our micro-experiment results. Choice of algorithm and motherset are constrained to be discrete factors, but our four problem dimensions can be described with numeric values; this should be especially useful in the case clusteredness as our construction criterion do not capture \emph{how} clustered or \emph{how} scattered the anomalies are.

We have already described measures of \textbf{clusteredness} and \textbf{feature irrelevance} in \cref{sec:method} that are continuous; as with clusteredness we will apply a log transform to the feature irrelevance ratio. \textbf{relative frequency} and \textbf{point difficulty} are both in the range $[0,1]$ and so as with AUC we will apply the logit transform to them. This gives us real valued representations of each problem dimension suitable for a linear model.

The simplest fixed effect linear model to build would be to predict a metric given our four problem dimensions (abbreviated $rf,pd,cl,ir$), choice of motherset ($mset$) and choice of algorithm ($algo$).
$$metric \sim rf + pd + cl + ir + mset + algo$$

First we build this model for each metric, using our discrete construction factors and our real valued transformations and compare the models. Our metric for comparison is the $\hat{R^2}$ goodness-of-fit measure, which is inversely related to the mean squared error of the model, which is the figure of merit each of these models is trying to optimize.

An ANOVA test on each of these models provides a $t$-test for each variable in the model and an $F$-test on the model itself. To save space we do not report these individual values; the results are easily summarized as \emph{every} test has a $p$-value well below $0.001$.

\begin{table}
\resizebox{\textwidth}{!}{\begin{minipage}{0.5\textwidth}
\tbl{$\hat{R^2}$ of Linear Regression by Metric and Variable Type\label{tbl:facvval}}{
\begin{tabular}{|r||c|c|}
\hline
&\textbf{$\text{logit}(AUC)$}&\textbf{$\log(LIFT)$}\\
\hline\hline
Discrete Variables&0.4910&0.6251\\
Real Variables&0.5019&0.6382\\
\hline
\end{tabular}}
\end{minipage}}
\end{table}

\begin{table}
\resizebox{\textwidth}{!}{\begin{minipage}{0.55\textwidth}
\tbl{Problem Dimension Coefficients by Metric\label{tbl:basecoef}}{
\begin{tabular}{|r||c|c|}
\hline
&\textbf{$\text{logit}(AUC)$}&\textbf{$\log(LIFT)$}\\
\hline\hline
$\text{logit}(\text{relative frequency})$&-0.1994&-0.3527\\
$\text{logit}(\text{point difficulty})$&-0.3209&-0.2014\\
$\text{clusteredness}$&-0.1255&-0.2141\\
$\log(\text{feature irrelevance})$&-0.2962&-0.1998\\
\hline
\end{tabular}}
\end{minipage}}
\end{table}


 \cref{tbl:facvval} shows the results of these first four models. The difference between using our construction factors and our real-valued problem dimensions is not large, but the real-valued variables do provide a better fit, are computationally less expensive for mixed effect models, and are easier to interpret as single coefficients, so we will use them in the remainder of our models.
 
 \cref{tbl:basecoef} shows the coefficients for each problem dimension in our base model. Note that they are all negative, which means each of them tends to make the benchmark more difficult. However, the scale of these coefficients are not comparable as the original variables are not at comparable scales.
 
To better evaluate the importance of each variable on predicting our metrics we construct simpler models that exclude one of the variables from the model and then measure the difference in $\hat{R^2}$ measures relative to our base model. We also construct a model without all four problem dimension variables to measure the impact of all problem dimensions in aggregate. \cref{tbl:rsq} shows these results. Boldfaced items are those with greater $\hat{R^2}$ loss than the algorithm variable which suggests that the variable is \emph{more important to your final outcome than your choice of algorithm is}.


\begin{table}
\resizebox{\textwidth}{!}{\begin{minipage}{1.15\textwidth}
\tbl{Changes in $\hat{R^2}$ When Different Variables Are Missing\label{tbl:rsq}}{
\begin{tabular}{|r||c|c|c|c|}
\hline
&\textbf{$R^2 (\text{logit}(AUC))$}&\textbf{$R^2 \text{loss} (\text{logit}(AUC))$}&\textbf{$R^2 (\log(LIFT))$}&\textbf{$R^2 \text{loss} (\log(LIFT))$}\\
\hline\hline
All Variables&0.5019&--&0.6382&--\\
w/o Algorithm&0.4512&0.0507&0.6013&0.0369\\
w/o Motherset&\textbf{0.2617}&\textbf{0.2403}&\textbf{0.5190}&\textbf{0.1192}\\
w/o All Problem Dimensions&\textbf{0.3221}&\textbf{0.1799}&\textbf{0.2359}&\textbf{0.4022}\\
-- w/o Relative Frequency&\textbf{0.4311}&\textbf{0.0709}&\textbf{0.4108}&\textbf{0.2274}\\
-- w/o Point Difficulty&0.4742&0.0277&0.6264&0.0117\\
-- w/o Clusteredness&0.4909&0.0111&\textbf{0.6004}&\textbf{0.0378}\\
-- w/o Feature Irrelevance&0.4831&0.0189&0.6279&0.0103\\
\hline
\end{tabular}}
\end{minipage}}
\end{table}

This suggests that the most important variable for predicting results is the choice of motherset, more important even than all four problem dimensions in aggregate, which has the second largest impact. Relative frequency alone has the third largest impact. Choice of algorithm is more important than each of the three remaining problem dimensions individually, but it may be surprising to see it so far down the list.

So far the $\hat{R^2}$ of our best models are 0.5019 and 0.6382 which are respectable measures but leave a lot of variance unexplained. We constructed several mixed effect models to see if we could better explain our results with the same data. Our best models kept our problem dimensions as fixed effects and treated choice of algorithm and motherset as random effect groups. Each member of each random effect group models its interaction with the fixed effects. Additionally, choice of motherset also models its own interaction with choice of algorithm.
$$metric \sim (rf + pd + cl + ir | algo) + (rf + pd + cl + ir + algo | mset)$$

\begin{figure}
\caption{}\label{fig:arco}
\includegraphics[width=\textwidth]{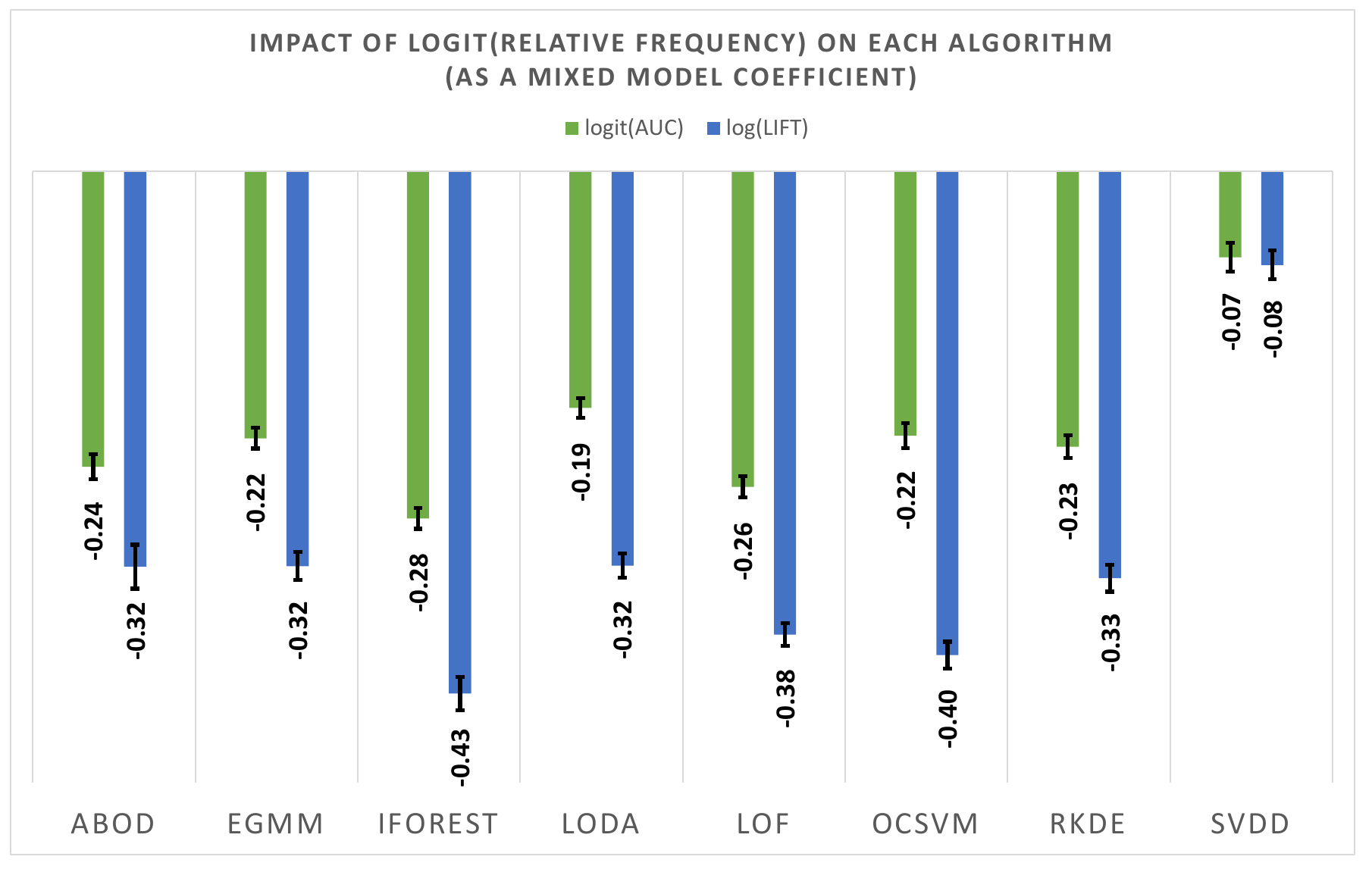}
\caption{}\label{fig:pdco}
\includegraphics[width=\textwidth]{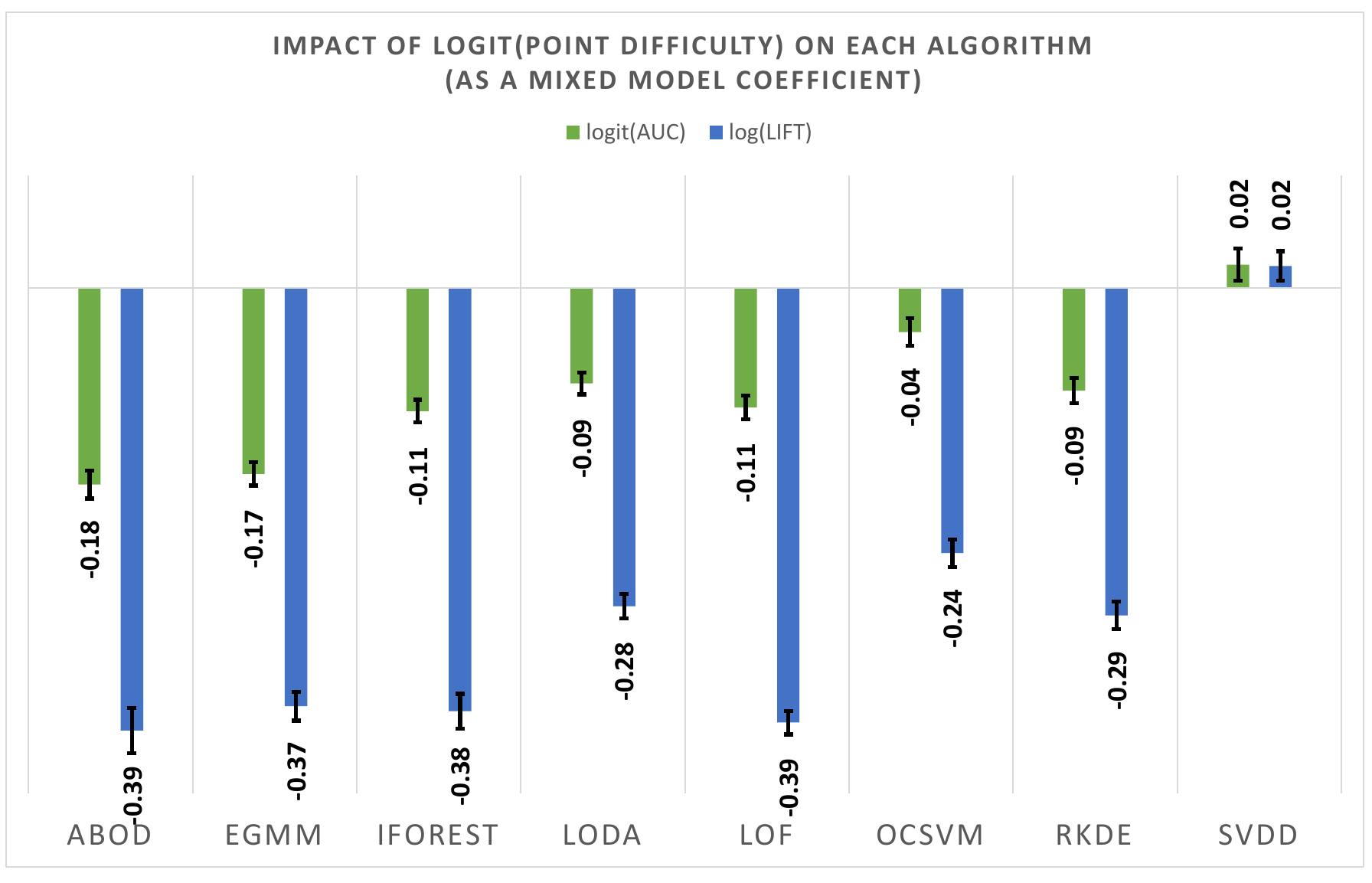}
\end{figure}

\begin{figure}
\caption{}\label{fig:clco}
\includegraphics[width=\textwidth]{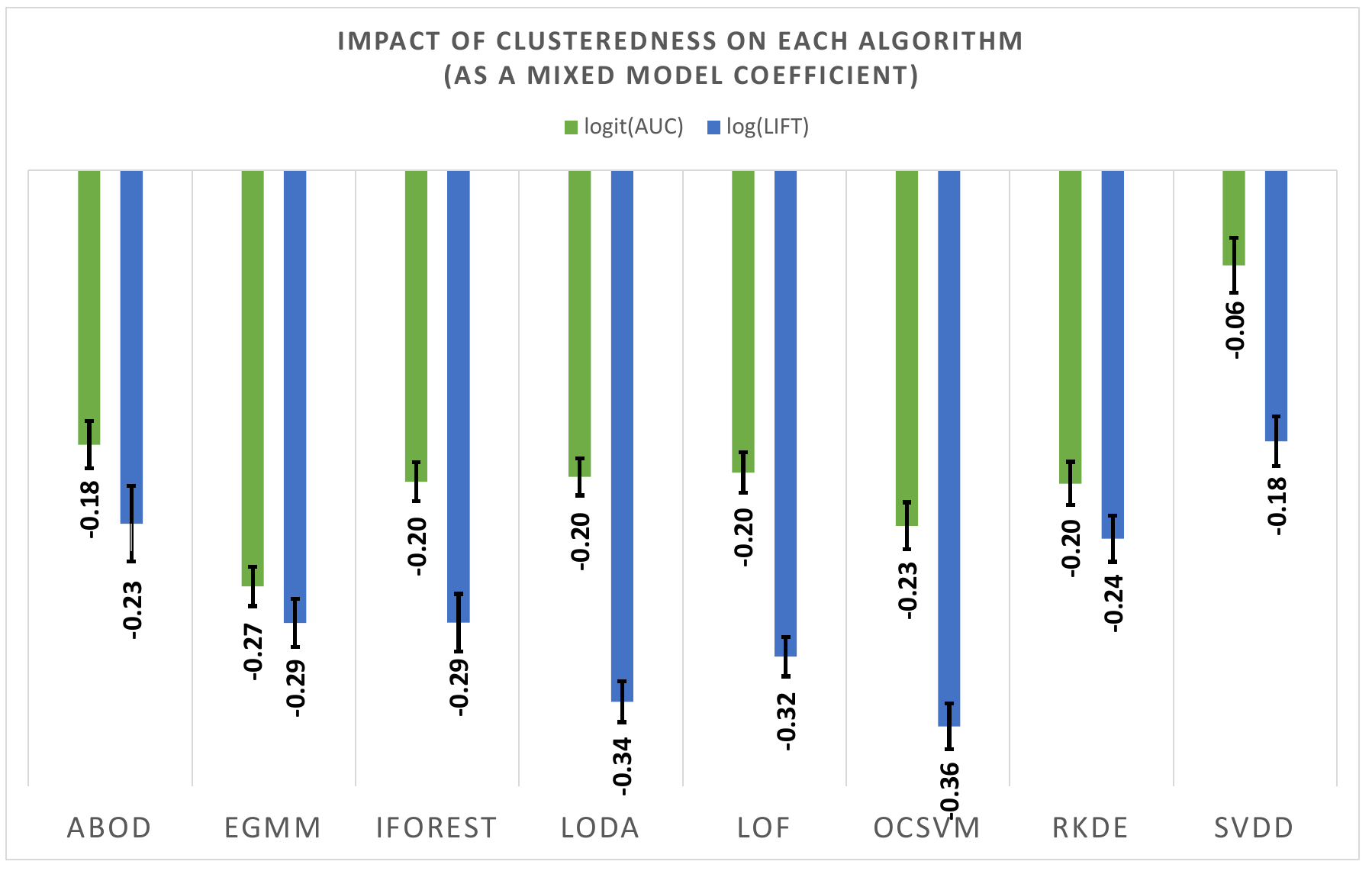}
\caption{}\label{fig:irco}
\includegraphics[width=\textwidth]{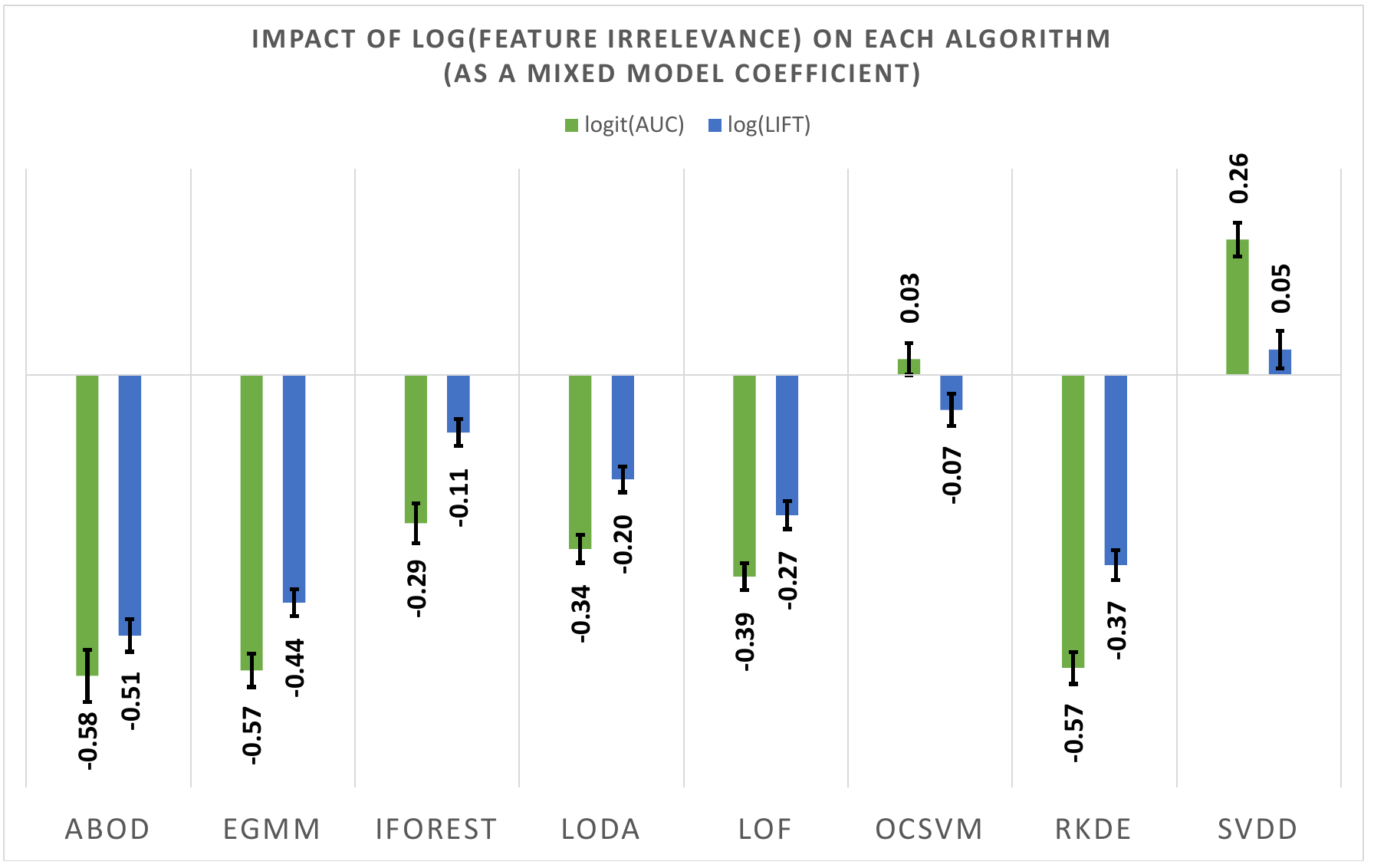}
\end{figure}

These models yield $\hat{R^2}$ measures of 0.7299 (for logit(AUC)) and 0.8009 (for log(LIFT)), a strong indication that the variables we are considering are adequate to explain micro-experiment outcomes. The mixed effects models also provide coefficients measuring the effect of each problem dimension on each algorithm. Similar to \cref{fig:ar,fig:pd,fig:cl,fig:ir}, \cref{fig:arco,fig:pdco,fig:clco,fig:irco} show these coefficients across each problem dimension.

Notably, there are some positive coefficients for the svdd and ocsvm algorithms. Given that these are the poorest performing algorithms overall, we urge caution in drawing conclusions about this. Further, we observe that these coefficients are only a part of a more complex model and do not tell the whole story; but they do offer some insight into the strengths and weaknesses of each algorithm.

\subsection{Further Comparison of Algorithms}\label{sec:dense}

So far Isolation Forest has been the top performing algorithm. \cref{fig:irco} would suggest that relative to the other competitive algorithms, Isolation Forest is not as negatively affected by the introduction of irrelevant features. This is in contrast with the author's own claims about Isolation Forest such as those in \cite{Liu:10} and \cite{inne} where they claim that the algorithm might fail in a larger feature space where only a small subset of those features are germane to the task. We find that this is not true, at least not relative to the other algorithms in our study, and to show it we recreate \cref{tbl:meanalgo} as \cref{tbl:noir,tbl:hiir}, considering only benchmarks constructed at feature irrelevance levels \textbf{ir-0} (No added noise features) and \textbf{ir-3} (many added noise features) respectively.

\begin{table}
\resizebox{\textwidth}{!}{\begin{minipage}{0.5\textwidth}
\tbl{Mean Performance by Algorithm and Metric with No Irrelevant Features\label{tbl:noir}}{
\begin{tabular}{|r||c|c|}
\hline
&\textbf{Mean $\text{logit}(AUC)$}&\textbf{Mean $\log(LIFT)$}\\
\hline\hline
abod&1.3706&\textbf{1.2792}\\
egmm&1.3522&1.2120\\
iforest&1.3145&1.1690\\
loda&1.1020&1.0235\\
lof&1.1800&1.0958\\
ocsvm&0.5654&0.9077\\
rkde&\textbf{1.4256}&1.1747\\
svdd&-0.0118&0.2290\\
\hline
\end{tabular}}
\end{minipage}}
\end{table}

\begin{table}
\resizebox{\textwidth}{!}{\begin{minipage}{0.5\textwidth}
\tbl{Mean Performance by Algorithm and Metric with Many Irrelevant Features\label{tbl:hiir}}{
\begin{tabular}{|r||c|c|}
\hline
&\textbf{Mean $\text{logit}(AUC)$}&\textbf{Mean $\log(LIFT)$}\\
\hline\hline
abod&0.4424&0.4612\\
egmm&0.4561&0.5069\\
iforest&\textbf{0.8117}&\textbf{1.0075}\\
loda&0.5803&0.7722\\
lof&0.6199&0.7187\\
ocsvm&0.5752&0.8527\\
rkde&0.5477&0.6716\\
svdd&0.3366&0.3580\\
\hline
\end{tabular}}
\end{minipage}}
\end{table}


\cref{tbl:noir} Shows an enormous improvement in the performance of egmm and rkde (The two density based methods) and also in abod. These were the algorithms most negatively affected by irrelevant features according to \cref{fig:irco}. By contrast in \cref{tbl:hiir} when only considering benchmarks with a large portion of irrelevant features, Isolation Forest only broadens its lead, outperforming all other algorithms by an even larger margin. This is a novel observation in itself, but it is also worth noting that if we had not included irrelevant features in this study, our overall comparison of algorithms would  be very different.

Similarly, \cref{tbl:hicl} shows the mean performance of each algorithm when only considering benchmarks with highly clustered anomalies; abod is the clear winner in this comparison. Also of note is that the other nearest-neighbor method, lof, surpasses most other algorithms too. Both of these algorithms account for local density in some way, and a possible explanation this is a tangible advantage when working with clustered anomalies. Again this also suggests that benchmark creation strategy can have a large impact on any final comparison of algorithms.

The latter point is important to recognize; imagine we had an interest in demonstrating the superiority of a particular algorithm. Keeping our methodology otherwise intact, we could use motherset selection to affect the final outcome. \cref{tbl:nefarious} shows the mean performance of each algorithm when only using the mothersets \textbf{abalone,particle,wine} and \textbf{yearp}. Loda and rkde benefit from this selection the most, and reporting one evaluation metric and not the other would allow you to declare whichever algorithm you preferred as the winner. While we don't suspect any such dishonest behavior abounds in the literature, this table provides further evidence of our assertion that selection of motherset and choice of evaluation metric are of great importance; most published work on anomaly detection do not address how these two factors might impact their reported results.


\begin{table}
\resizebox{\textwidth}{!}{\begin{minipage}{0.5\textwidth}
\tbl{Mean Performance by Algorithm and Metric When Clusteredness is Greater Than 0.25\label{tbl:hicl}}{
\begin{tabular}{|r||c|c|}
\hline
&\textbf{Mean $\text{logit}(AUC)$}&\textbf{Mean $\log(LIFT)$}\\
\hline\hline
abod&\textbf{1.0305}&\textbf{0.7585}\\
egmm&0.7199&0.4834\\
iforest&0.7405&0.5075\\
loda&0.5689&0.4225\\
lof&0.7623&0.5336\\
ocsvm&0.0385&0.2316\\
rkde&0.7179&0.5387\\
svdd&0.0327&0.0639\\
\hline
\end{tabular}}
\end{minipage}}
\end{table}

\begin{table}
\resizebox{\textwidth}{!}{\begin{minipage}{0.5\textwidth}
\tbl{Mean Performance by Algorithm and Metric with Mothersets Selected for Dubious Reasons\label{tbl:nefarious}}{
\begin{tabular}{|r||c|c|}
\hline
&\textbf{Mean $\text{logit}(AUC)$}&\textbf{Mean $\log(LIFT)$}\\
\hline\hline
abod&0.6094&0.6071\\
egmm&0.5826&0.6285\\
iforest&0.7675&0.8903\\
loda&0.7981&\textbf{1.0060}\\
lof&0.5999&0.7340\\
ocsvm&0.5121&0.8080\\
rkde&\textbf{0.8554}&0.9222\\
svdd&0.1764&0.3164\\
\hline
\end{tabular}}
\end{minipage}}
\end{table}

\subsection{A Trivial Solution}\label{sec:trivoracle}

By eliminating benchmarks based on hypothesis testing we attempted to account for benchmarks that are too difficult or otherwise unrealistic in composition, but an additional concern of ours is that some benchmarks might be trivially easy. It is often the case in anomaly detection literature that reported results are highly accurate, such as in \cite{loda,inne,Kriegel:2009,rajasegarar2010centered,Amer:2013} whereas the mean accuracy on our corpus of benchmarks is much lower. In particular, consider the work presented in \cite{inne}; the authors share the parameterization of each algorithm on each benchmark and praise the algorithm \textbf{iNNe} for sometimes performing well with parameter $\psi=2$. However, an understanding of the iNNe algorithm and the implication of that parameter choice will reveal that the algorithm is doing little more than approximating the distance of each point from the mean of the data. While we acknowledge that this particular work is a smaller workshop publication, it should serve as a warning that benchmarks for which all algorithms perform well might be benchmarks that can be trivially solved and their inclusion in reported results might not be helpful.

To investigate this phenomenon further, we ran a trivial algorithm against our corpus of benchmarks. The algorithm simply computes the arithmetic mean of the data and assigns an outlier score to each point based on its euclidean distance from that mean. We use the performance of this algorithm to normalize the performance of the other non-trivial algorithms. For both AUC and AP we compute the ratio of a given algorithm's performance over the performance of the trivial algorithm. As with logit$(AUC)$ and $\log(LIFT)$ we then take the natural log of these ratios.
$$\log\left(\frac{AUC_{\text{non-trivial}}}{AUC_{\text{trivial}}}\right), \log\left(\frac{AP_{\text{non-trivial}}}{AP_{\text{trivial}}}\right) $$
In the case of AP, this is very similar to our previous transformed metric, except instead of normalizing against random expectation, we are normalizing against the performance of a trivial solution. Under this metric, achieving a perfect score on a benchmark that is also perfectly or almost-perfectly solved by a trivial solution is not given much merit, while a lower score that is a significant improvement on trivial performance is given much more credit. \cref{tbl:triv} shows the mean performance of each algorithm under these new metrics.

\begin{table}
\resizebox{\textwidth}{!}{\begin{minipage}{0.63\textwidth}
\tbl{Mean Performance by Algorithm and Metric (Normalized by a Trivial Solution)\label{tbl:triv}}{
\begin{tabular}{|r||c|c|}
\hline
&\textbf{Mean $\log\left(\frac{AUC_{\text{non-trivial}}}{AUC_{\text{trivial}}}\right)$}&\textbf{Mean $\log\left(\frac{AP_{\text{non-trivial}}}{AP_{\text{trivial}}}\right)$}\\
\hline\hline
abod&0.0654&0.1193\\
egmm&0.0774&0.1265\\
iforest&\textbf{0.1006}&\textbf{0.3102}\\
loda&0.0578&0.1340\\
lof&0.0723&0.1513\\
ocsvm&-0.1004&0.0792\\
rkde&0.0707&0.1316\\
svdd&-0.1817&-0.5010\\
\hline
\end{tabular}}
\end{minipage}}
\end{table}

Again Isolation Forest is the best overall. Interestingly, the classic LOF algorithm outperforms most others in this context. Most of these scores demonstrate only a marginal improvement on a trivial solution.

\section{Conclusions and Recommendations}\label{sec:discuss}

Given the evidence provided, we feel confident making several conclusions and recommendations. We acknowledge that many conclusions are only one possible explanation.

\cref{sec:hypo} reveals that it is not uncommon for benchmarks created by methods common in the literature to be of a quality such that many algorithms cannot distinguish themselves from a random ranking in a statistical hypothesis test. In \cref{sec:trivoracle} we observe that many experiments in literature are in danger of being too easy, not too hard, and so we do not believe that many reported findings are statistically insignificant, but we do feel it is poor form to not address the matter.

Due to the statistical hypothesis tests we performed, we recommend against creating benchmarks with very high point difficulty (\cref{tbl:pdfail}) and we urge caution when selecting data sets to create benchmarks from. Based on particularly high failure rates reported in \cref{tbl:msetfail} we recommend against using the yearp,yeast and letter.rec datasets in future experiments and it would be beneficial to further validate commonly used datasets.

More generally, \cref{tbl:origfail} shows that benchmarks created from binary mothersets have a lower failure rate, suggesting that beginning with a binary classification dataset might be preferable in future work as it is more likely to produce statistically significant results.

We also recommend against an uncontrolled relative frequency, but \cref{tbl:arfail} does not provide evidence to support this. The confidence intervals used in our hypothesis tests are themselves functions of the relative frequency and so the failure rates in this context are not as informative. However, the fact that the failure rates at the uncontrolled level were lower than average does provide some evidence that anomaly detection algorithms in general are capable of picking up some amount of signal (albeit weak) as long as there is some imbalance in the classes. This reinforces results published in \cite{Liu:08} where results were positive even the reported anomaly rate was high.

Benchmarks with low relative frequency had a high failure rate, but \cref{fig:ar} suggests that the remainder of those benchmarks were also relatively easy compared to other benchmarks. This can be explained by what should be intuitively obvious: that benchmarks with very few anomalies are going to see much higher variance in their results, meaning more outstanding successes and more abject failures.

Based on \cref{fig:mset} we recommend against using synthetic datasets in general, and further \cref{tbl:rsq} suggests that the biggest factor impacting experimental results is the selection of motherset. \cref{tbl:nefarious} further demonstrates that it is possible to quietly tailor an experiment to favor a particular algorithm. While we doubt any such dishonesty exists in the literature, most reported results are derived from a somewhat arbitrary list of datasets. It is possible that some positive findings in literature are due to a lucky selection of datasets. The best protection against this is to simply use as many sources as is reasonable, or to otherwise justify the selection of datasets used in experiments.

\cref{fig:ar,fig:pd,fig:cl,fig:ir} and \cref{tbl:basecoef} suggest that our defined problems dimensions all have an impact on experimental results. The reported $\hat{R^2}$ of our mixed models in \cref{sec:basic} suggest that choice of motherset, choice of algorithm and our proposed problem dimensions are capable of predicting experimental results with good accuracy. Based on this we are able to recommend using our methodology (or something appropriately similar) for controlling and measuring these problem dimensions. We encourage further work that focuses on specific contexts that can be defined by these problem dimensions, especially if it maps these contexts to real-world applications.

Because Isolation Forest performed best on average and because it has very good runtime properties, we recommend it for general use. However, we also recommend that context should impact your choice of algorithm. Based on \cref{fig:irco} and \cref{tbl:noir,tbl:hiir} we observe that if you are confident in your feature space, probability density estimates and ABOD outperform Isolation Forest. In large feature spaces of unknown quality we recommend Isolation Forest.

Based on \cref{fig:clco} and \cref{tbl:hicl} we recommend ABOD or LOF for highly clustered anomalies.

We also observe that Isolation Forest and LODA scale well to large data sets while the other algorithms do not. Similarly, density estimating methods such as rkde and egmm do not scale well to a large number of features.

In general, SVDD and OC-SVM performed very poorly compared to other algorithms. We made our best effort to parameterize them well. \cref{appendix:parameters} details our parameterization and what other works they are based on. However, we can not conclude that these are poor algorithms, but rather that they are difficult to parameterize correctly and we were unable to get them to perform competitively with the other algorithms in this study. However, difficulty of use would be a reason not to recommend an algorithm.

Among the other algorithms we point out that the difference in performance among them was not very large, so while we do observe that Isolation Forest did the best overall, we point out that the field of anomaly detection algorithms are more or less solving benchmarks with the same efficacy. Experimental design and understanding the impact of different real world contexts seem to be of more importance given the evidence in this study. \cref{sec:trivoracle} provides some evidence that over-positive results reported in literature are not particularly helpful measuring progress. Incremental improvements in non-standardized environments demonstrate that a particular algorithm is effective, but they do not demonstrate any particular breakthroughs in the field. It is our opinion that on average most algorithms are roughly measuring the same quantity and producing the same results. Worse, \cref{tbl:triv} might suggest that often algorithms are only marginally better than a trivial solution.

It is also our opinion that it is a matter of optics that AUC is the most commonly reported metric, and that most reported results are overwhelmingly positive. Well-mastered benchmarks do not allow any room for improvement and a lack of standard benchmarks do not enable the recognition of true breakthroughs in the field. We recommend our own corpus of benchmarks as the \emph{beginning} of a standardized test bed. The corpus and the software that produced it can be found at:

\ 

\url{http://ir.library.oregonstate.edu/xmlui/handle/1957/59114}

\ 

We welcome future contributions in this area as well as criticisms and refinements of our existing corpus.

\bibliographystyle{ACM-Reference-Format-Journals}
\bibliography{bencon}


\begin{thebibliography}{00}


\ifx \showCODEN    \undefined \def \showCODEN     #1{\unskip}     \fi
\ifx \showDOI      \undefined \def \showDOI       #1{{\tt DOI:}\penalty0{#1}\ }
  \fi
\ifx \showISBNx    \undefined \def \showISBNx     #1{\unskip}     \fi
\ifx \showISBNxiii \undefined \def \showISBNxiii  #1{\unskip}     \fi
\ifx \showISSN     \undefined \def \showISSN      #1{\unskip}     \fi
\ifx \showLCCN     \undefined \def \showLCCN      #1{\unskip}     \fi
\ifx \shownote     \undefined \def \shownote      #1{#1}          \fi
\ifx \showarticletitle \undefined \def \showarticletitle #1{#1}   \fi
\ifx \showURL      \undefined \def \showURL       #1{#1}          \fi

\bibitem[\protect\citeauthoryear{Aggarwal, Danda, Gupta, Gehlot,
  et~al\mbox{.}}{Aggarwal et~al\mbox{.}}{2009}]%
        {aggarwal2009}
{Bharat~B Aggarwal}, {Divya Danda}, {Shan Gupta}, {Prashasnika Gehlot}, {and}
  {others}. 2009.
\newblock \showarticletitle{Models For Prevention and Treatment of Cancer:
  Problems vs Promises}.
\newblock {\em Biochemical pharmacology\/} {78}, 9 (2009), 1083.
\newblock


\bibitem[\protect\citeauthoryear{Alzghoul and Lofstrand}{Alzghoul and
  Lofstrand}{2011}]%
        {Alzghoul:2011}
{Ahmad Alzghoul} {and} {Magnus Lofstrand}. 2011.
\newblock \showarticletitle{Increasing Availability of Industrial Systems
  Through Data Stream Mining}.
\newblock {\em Computers \& Industrial Engineering\/} {60}, 2 (2011), 195 --
  205.
\newblock
\showISSN{0360-8352}
\showDOI{%
\url{http://dx.doi.org/10.1016/j.cie.2010.10.008}}


\bibitem[\protect\citeauthoryear{Amer, Goldstein, and Abdennadher}{Amer
  et~al\mbox{.}}{2013}]%
        {Amer:2013}
{Mennatallah Amer}, {Markus Goldstein}, {and} {Slim Abdennadher}. 2013.
\newblock \showarticletitle{Enhancing One-class Support Vector Machines for
  Unsupervised Anomaly Detection}. In {\em Proceedings of the ACM SIGKDD
  Workshop on Outlier Detection and Description} {\em (ODD '13)}. ACM, New
  York, NY, USA, 8--15.
\newblock
\showISBNx{978-1-4503-2335-2}
\showDOI{%
\url{http://dx.doi.org/10.1145/2500853.2500857}}


\bibitem[\protect\citeauthoryear{Bache and Lichman}{Bache and Lichman}{2013}]%
        {uci}
{K. Bache} {and} {M. Lichman}. 2013.
\newblock {UCI} Machine Learning Repository.
\newblock   (2013).
\newblock
\showURL{%
\url{http://archive.ics.uci.edu/ml}}


\bibitem[\protect\citeauthoryear{Bandaragoda, Ting, Albrecht, Liu, and
  Wells}{Bandaragoda et~al\mbox{.}}{2014}]%
        {inne}
{Tharindu~R Bandaragoda}, {Kai~Ming Ting}, {David Albrecht}, {Fei~Tony Liu},
  {and} {Jason~R Wells}. 2014.
\newblock \showarticletitle{Efficient Anomaly Detection by Isolation Using
  Nearest Neighbour Ensemble}. In {\em Data Mining Workshop (ICDMW), 2014 IEEE
  International Conference on}. IEEE, 698--705.
\newblock


\bibitem[\protect\citeauthoryear{Bestgen}{Bestgen}{2015}]%
        {exactAP}
{Yves Bestgen}. 2015.
\newblock \showarticletitle{Exact expected average precision of the random
  baseline for system evaluation}.
\newblock {\em The Prague Bulletin of Mathematical Linguistics\/} {103}, 1
  (2015), 131--138.
\newblock


\bibitem[\protect\citeauthoryear{Breiman}{Breiman}{2001}]%
        {Breiman}
{Leo Breiman}. 2001.
\newblock \showarticletitle{Random Forests}.
\newblock {\em Machine Learning\/} {45}, 1 (2001), 5--32.
\newblock


\bibitem[\protect\citeauthoryear{Breunig, Kriegel, Raymond T.~Ng, and
  Sander}{Breunig et~al\mbox{.}}{2000}]%
        {Breunig:2000}
{M. Breunig}, {H-P. Kriegel}, {R.~T. Raymond T.~Ng}, {and} {J. Sander}. 2000.
\newblock \showarticletitle{{LOF:} Identifying Density-Based Local Outliers}.
\newblock {\em ACM SIGMOD Record\/} (2000), 93--104.
\newblock


\bibitem[\protect\citeauthoryear{Bridges, Collins, Ferragut, Laska, and
  Sullivan}{Bridges et~al\mbox{.}}{2014}]%
        {bcfls-mladsgd-2014}
{Robert~A. Bridges}, {John Collins}, {Eric Ferragut}, {Jason Laska}, {and}
  {Blair~D. Sullivan}. 2014.
\newblock {\em Multi-Level Anomaly Detection on Streaming Graph Data}.
\newblock {T}echnical {R}eport 1410.4355v1. arXiv.
\newblock


\bibitem[\protect\citeauthoryear{Chang and Lin.}{Chang and Lin.}{2011}]%
        {chang:2011}
{C.-C. Chang} {and} {C.-J. Lin.} 2011.
\newblock \showarticletitle{{LIBSVM}: {A} library for support vector machines}.
\newblock {\em ACM Transactions on Intelligent Systems and Technology\/}  {2}
  (2011), 27:1--27:27.
\newblock


\bibitem[\protect\citeauthoryear{Cortez, Cerdeira, Almeida, Matos, and
  Reis}{Cortez et~al\mbox{.}}{2009}]%
        {cortez2009}
{Paulo Cortez}, {Ant{\'o}nio Cerdeira}, {Fernando Almeida}, {Telmo Matos},
  {and} {Jos{\'e} Reis}. 2009.
\newblock \showarticletitle{Modeling Wine Preferences by Data Mining from
  Physicochemical Properties}.
\newblock {\em Decision Support Systems\/} {47}, 4 (2009), 547--553.
\newblock


\bibitem[\protect\citeauthoryear{Das, Schneider, and Neill}{Das
  et~al\mbox{.}}{2008}]%
        {das2008}
{Kaustav Das}, {Jeff Schneider}, {and} {Daniel~B Neill}. 2008.
\newblock \showarticletitle{Anomaly Pattern Detection in Categorical Datasets}.
  In {\em Proceedings of the 14th ACM SIGKDD International Conference on
  Knowledge Discovery and Data Mining}. ACM, 169--176.
\newblock


\bibitem[\protect\citeauthoryear{Denning}{Denning}{1987}]%
        {denning1987}
{Dorothy~E Denning}. 1987.
\newblock \showarticletitle{An Intrusion-Detection Model}.
\newblock {\em IEEE Transactions on Software Engineering\/} 2 (1987), 222--232.
\newblock


\bibitem[\protect\citeauthoryear{Dereszynski and Dietterich}{Dereszynski and
  Dietterich}{2011}]%
        {Dereszynski2011}
{Ethan Dereszynski} {and} {Thomas~G Dietterich}. 2011.
\newblock \showarticletitle{Spatiotemporal Models for Anomaly Detection in
  Dynamic Environmental Monitoring Campaigns}.
\newblock {\em ACM Transactions on Sensor Networks\/} {8}, 1 (2011), 3:1--3:26.
\newblock


\bibitem[\protect\citeauthoryear{Elkan}{Elkan}{1999}]%
        {kdd99}
{C. Elkan}. 1999.
\newblock Results of the KDD'99 Classifier Learning Contest.
\newblock   (1999).
\newblock
\showURL{%
\url{http://www.cs.ucsd.edu/users/elkan/clresults.html}}


\bibitem[\protect\citeauthoryear{Emmott, Das, Dietterich, Fern, and
  Wong}{Emmott et~al\mbox{.}}{2015}]%
        {bencon2015}
{Andrew Emmott}, {Shubhomoy Das}, {Thomas~G. Dietterich}, {Alan Fern}, {and}
  {Weng{-}Keen Wong}. 2015.
\newblock \showarticletitle{Systematic Construction of Anomaly Detection
  Benchmarks from Real Data}.
\newblock {\em CoRR\/}  {abs/1503.01158} (2015).
\newblock
\showURL{%
\url{http://arxiv.org/abs/1503.01158}}


\bibitem[\protect\citeauthoryear{Emmott, Das, Dietterich, Fern, and
  Wong}{Emmott et~al\mbox{.}}{2013}]%
        {odd2013}
{Andrew~F Emmott}, {Shubhomoy Das}, {Thomas Dietterich}, {Alan Fern}, {and}
  {Weng-Keen Wong}. 2013.
\newblock \showarticletitle{Systematic construction of anomaly detection
  benchmarks from real data}. In {\em Proceedings of the ACM SIGKDD workshop on
  outlier detection and description}. ACM, 16--21.
\newblock


\bibitem[\protect\citeauthoryear{Glasser and Lindauer}{Glasser and
  Lindauer}{2013}]%
        {glasser2013}
{Joshua Glasser} {and} {Brian Lindauer}. 2013.
\newblock \showarticletitle{Bridging the Gap: A Pragmatic Approach to
  Generating Insider Threat Data}. In {\em 2013 IEEE Security and Privacy
  Workshops}. IEEE Press, 98--104.
\newblock


\bibitem[\protect\citeauthoryear{Greensmith, Twycross, and Aickelin}{Greensmith
  et~al\mbox{.}}{2006}]%
        {greensmith:06}
{Julie Greensmith}, {Jamie Twycross}, {and} {Uwe Aickelin}. 2006.
\newblock \showarticletitle{Dendritic Cells for Anomaly Detection}. In {\em
  IEEE Congress on Evolutionary Computation}. IEEE, 664--671.
\newblock


\bibitem[\protect\citeauthoryear{Hawkins}{Hawkins}{1980}]%
        {Hawkins1980}
{D.M. Hawkins}. 1980.
\newblock {\em Identification of Outliers}.
\newblock Chapman and Hall.
\newblock
\showISBNx{9780412219009}
\showLCCN{81192025}
\showURL{%
\url{http://books.google.com/books?id=fb0OAAAAQAAJ}}


\bibitem[\protect\citeauthoryear{Huang, Mehrotra, and Mohan}{Huang
  et~al\mbox{.}}{2013}]%
        {huang2013}
{Huaming Huang}, {Kishan Mehrotra}, {and} {Chilukuri~K Mohan}. 2013.
\newblock \showarticletitle{An Online Anomalous Time Series Detection Algorithm
  for Univariate Data Streams}.
\newblock In {\em Recent Trends in Applied Artificial Intelligence}. Springer,
  151--160.
\newblock


\bibitem[\protect\citeauthoryear{Jaakkola and Haussler}{Jaakkola and
  Haussler}{1999}]%
        {klr1}
{Tommi Jaakkola} {and} {David Haussler}. 1999.
\newblock \showarticletitle{Probabilistic Kernel Regression Models}. In {\em
  Proceedings of the 1999 Conference on AI and Statistics}, Vol. 126. San
  Mateo, CA, 00--04.
\newblock


\bibitem[\protect\citeauthoryear{Kakde, Chaudhuri, Kong, Jahja, Jiang, and
  Silva}{Kakde et~al\mbox{.}}{2016}]%
        {kakde2016}
{Deovrat Kakde}, {Arin Chaudhuri}, {Seunghyun Kong}, {Maria Jahja}, {Hansi
  Jiang}, {and} {Jorge Silva}. 2016.
\newblock \showarticletitle{Peak Criterion for Choosing Gaussian Kernel
  Bandwidth in Support Vector Data Description}.
\newblock {\em arXiv preprint arXiv:1602.05257\/} (2016).
\newblock


\bibitem[\protect\citeauthoryear{Keerthi, Duan, Shevade, and Poo}{Keerthi
  et~al\mbox{.}}{2005}]%
        {klr3}
{S.S. Keerthi}, {K.B. Duan}, {S.K. Shevade}, {and} {A.N. Poo}. 2005.
\newblock \showarticletitle{A Fast Dual Algorithm for Kernel Logistic
  Regression}.
\newblock {\em Machine Learning\/} {61}, 1-3 (2005), 151--165.
\newblock
\showISSN{0885-6125}
\showDOI{%
\url{http://dx.doi.org/10.1007/s10994-005-0768-5}}


\bibitem[\protect\citeauthoryear{Kim and Scott}{Kim and Scott}{2008}]%
        {kim:08}
{Joo~Seuk Kim} {and} {C. Scott}. 2008.
\newblock \showarticletitle{Robust Kernel Density Estimation}. In {\em IEEE
  International Conference on Acoustics, Speech and Signal Processing}.
  3381--3384.
\newblock
\showISSN{1520-6149}
\showDOI{%
\url{http://dx.doi.org/10.1109/ICASSP.2008.4518376}}


\bibitem[\protect\citeauthoryear{Kriegel, Kr\"{o}ger, Schubert, and
  Zimek}{Kriegel et~al\mbox{.}}{2009}]%
        {Kriegel:2009}
{Hans-Peter Kriegel}, {Peer Kr\"{o}ger}, {Erich Schubert}, {and} {Arthur
  Zimek}. 2009.
\newblock \showarticletitle{LoOP: Local Outlier Probabilities}. In {\em
  Proceedings of the 18th ACM Conference on Information and Knowledge
  Management} {\em (CIKM '09)}. ACM, New York, NY, USA, 1649--1652.
\newblock
\showISBNx{978-1-60558-512-3}
\showDOI{%
\url{http://dx.doi.org/10.1145/1645953.1646195}}


\bibitem[\protect\citeauthoryear{Kriegel, Zimek, et~al\mbox{.}}{Kriegel
  et~al\mbox{.}}{2008}]%
        {abod}
{Hans-Peter Kriegel}, {Arthur Zimek}, {and} {others}. 2008.
\newblock \showarticletitle{Angle-based outlier detection in high-dimensional
  data}. In {\em Proceedings of the 14th ACM SIGKDD international conference on
  Knowledge discovery and data mining}. ACM, 444--452.
\newblock


\bibitem[\protect\citeauthoryear{Lane and Brodley}{Lane and Brodley}{1997}]%
        {lane1997sequence}
{Terran Lane} {and} {Carla~E Brodley}. 1997.
\newblock \showarticletitle{Sequence Matching and Learning in Anomaly Detection
  for Computer Security}. In {\em AAAI Workshop: AI Approaches to Fraud
  Detection and Risk Management}. 43--49.
\newblock


\bibitem[\protect\citeauthoryear{Lazarevic, Ertoz, Kumar, Ozgur, and
  Srivastava}{Lazarevic et~al\mbox{.}}{2003}]%
        {Lazarevic:2003}
{Ar Lazarevic}, {Levent Ertoz}, {Vipin Kumar}, {Aysel Ozgur}, {and} {Jaideep
  Srivastava}. 2003.
\newblock \showarticletitle{A comparative study of anomaly detection schemes in
  network intrusion detection}. In {\em In Proceedings of SIAM Conference on
  Data Mining}.
\newblock


\bibitem[\protect\citeauthoryear{Liu, Ting, and Zhou}{Liu
  et~al\mbox{.}}{2008}]%
        {Liu:08}
{F~T Liu}, {K~M Ting}, {and} {Z-H Zhou}. 2008.
\newblock \showarticletitle{Isolation Forest}. In {\em Proceedings of the IEEE
  International Conference on Data Mining}. 413--422.
\newblock


\bibitem[\protect\citeauthoryear{Liu, Ting, and Zhou}{Liu
  et~al\mbox{.}}{2010}]%
        {Liu:10}
{F~T Liu}, {K~M Ting}, {and} {Z-H Zhou}. 2010.
\newblock \showarticletitle{On Detecting Clustered Anomalies using
  {SCiForest}}. In {\em Machine Learning and Knowledge Discovery in Databases}.
  274--290.
\newblock


\bibitem[\protect\citeauthoryear{Mahoney and Chan}{Mahoney and Chan}{2003}]%
        {mahoney2003}
{Matthew~V Mahoney} {and} {Philip~K Chan}. 2003.
\newblock \showarticletitle{An Analysis of the 1999 DARPA/Lincoln Laboratory
  Evaluation Data for Network Anomaly Detection}. In {\em Recent Advances in
  Intrusion Detection}. Springer, 220--237.
\newblock


\bibitem[\protect\citeauthoryear{McHugh}{McHugh}{2000}]%
        {mchugh2000}
{John McHugh}. 2000.
\newblock \showarticletitle{Testing Intrusion Detection Systems: a Critique of
  the 1998 and 1999 DARPA Intrusion Detection System Evaluations as Performed
  by Lincoln Laboratory}.
\newblock {\em ACM transactions on Information and system Security\/} {3}, 4
  (2000), 262--294.
\newblock


\bibitem[\protect\citeauthoryear{Mei and Gul}{Mei and Gul}{2013}]%
        {mei2013}
{Qipei Mei} {and} {Mustafa Gul}. 2013.
\newblock \showarticletitle{An Improved Methodology for Anomaly Detection Based
  on Time Series Modeling}.
\newblock In {\em Topics in Dynamics of Civil Structures, Volume 4}. Springer,
  277--281.
\newblock


\bibitem[\protect\citeauthoryear{Otey, Ghoting, and Parthasarathy}{Otey
  et~al\mbox{.}}{2006}]%
        {otey2006}
{Matthew~Eric Otey}, {Amol Ghoting}, {and} {Srinivasan Parthasarathy}. 2006.
\newblock \showarticletitle{Fast Distributed Outlier Detection in
  Mixed-Attribute Data Sets}.
\newblock {\em Data Mining and Knowledge Discovery\/} {12}, 2-3 (2006),
  203--228.
\newblock


\bibitem[\protect\citeauthoryear{Pevn{\`y}}{Pevn{\`y}}{2016}]%
        {loda}
{Tom{\'a}{\v{s}} Pevn{\`y}}. 2016.
\newblock \showarticletitle{Loda: Lightweight on-line detector of anomalies}.
\newblock {\em Machine Learning\/} {102}, 2 (2016), 275--304.
\newblock


\bibitem[\protect\citeauthoryear{Pokrajac, Lazarevic, and Latecki}{Pokrajac
  et~al\mbox{.}}{2007}]%
        {Lazarevic:2007}
{D. Pokrajac}, {A. Lazarevic}, {and} {L.J. Latecki}. 2007.
\newblock \showarticletitle{Incremental Local Outlier Detection for Data
  Streams}. In {\em IEEE Symposium on Computational Intelligence and Data
  Mining}. 504--515.
\newblock
\showDOI{%
\url{http://dx.doi.org/10.1109/CIDM.2007.368917}}


\bibitem[\protect\citeauthoryear{Polat, Sahan, Kodaz, and Gunes}{Polat
  et~al\mbox{.}}{2005}]%
        {Polat:05}
{Kemal Polat}, {Seral Sahan}, {Halife Kodaz}, {and} {Salih Gunes}. 2005.
\newblock \showarticletitle{A New Classification Method for Breast Cancer
  Diagnosis: Feature Selection Artificial Immune Recognition System (FS-AIRS)}.
\newblock In {\em Advances in Natural Computation}, {Lipo Wang}, {Ke~Chen},
  {and} {YewSoon Ong} (Eds.). Lecture Notes in Computer Science, Vol. 3611.
  Springer Berlin Heidelberg, 830--838.
\newblock
\showISBNx{978-3-540-28325-6}
\showDOI{%
\url{http://dx.doi.org/10.1007/11539117_117}}


\bibitem[\protect\citeauthoryear{Portnoy, Eskin, and Stolfo}{Portnoy
  et~al\mbox{.}}{2001}]%
        {portnoy2001intrusion}
{Leonid Portnoy}, {Eleazar Eskin}, {and} {Sal Stolfo}. 2001.
\newblock \showarticletitle{Intrusion Detection with Unlabeled Data Using
  Clustering}. In {\em In Proceedings of ACM CSS Workshop on Data Mining
  Applied to Security (DMSA-2001}. Citeseer.
\newblock


\bibitem[\protect\citeauthoryear{Rajasegarar, Leckie, Bezdek, and
  Palaniswami}{Rajasegarar et~al\mbox{.}}{2010}]%
        {rajasegarar2010centered}
{Sutharshan Rajasegarar}, {Christopher Leckie}, {James~C Bezdek}, {and}
  {Marimuthu Palaniswami}. 2010.
\newblock \showarticletitle{Centered hyperspherical and hyperellipsoidal
  one-class support vector machines for anomaly detection in sensor networks}.
\newblock {\em Information Forensics and Security, IEEE Transactions on\/} {5},
  3 (2010), 518--533.
\newblock


\bibitem[\protect\citeauthoryear{Rocke and Woodruff}{Rocke and
  Woodruff}{1996}]%
        {mulcross}
{David~M. Rocke} {and} {David~L. Woodruff}. 1996.
\newblock \showarticletitle{Identification of Outliers in Multivariate Data}.
\newblock {\it J. Amer. Statist. Assoc.} {91}, 435 (1996), 1047--1061.
\newblock


\bibitem[\protect\citeauthoryear{Scholkopf, Platt, Shawe-taylor, Smola, and
  Williamson}{Scholkopf et~al\mbox{.}}{1999}]%
        {Scholkopf:99}
{B. Scholkopf}, {J.~C. Platt}, {J. Shawe-taylor}, {A.~J. Smola}, {and} {R.~C.
  Williamson}. 1999.
\newblock Estimating the Support of a High-Dimensional Distribution.
\newblock   (1999).
\newblock


\bibitem[\protect\citeauthoryear{Senator, Henry, Goldberg, Young, Rees, Pierce,
  Huang, Reardon, Lee, Koutra, et~al\mbox{.}}{Senator et~al\mbox{.}}{2013}]%
        {senator2013a}
{Ted~E Senator}, {G Henry}, {Alex~Memory Goldberg}, {William~T Young}, {Brad
  Rees}, {Robert Pierce}, {Daniel Huang}, {Matthew Reardon}, {Jay-Yoon Lee},
  {Danai Koutra}, {and} {others}. 2013.
\newblock \showarticletitle{Detecting Insider Threats in a Real Corporate
  Database of Computer Usage Activity}.
\newblock  (2013).
\newblock


\bibitem[\protect\citeauthoryear{Tax and Duin}{Tax and Duin}{2004}]%
        {Tax:2004}
{Tax} {and} {Duin}. 2004.
\newblock \showarticletitle{Support Vector Data Description}.
\newblock {\em Machine Learning\/}  {54} (2004), 45--66.
\newblock


\bibitem[\protect\citeauthoryear{Wagstaff, Lanza, Thompson, Dietterich, and
  Gilmore}{Wagstaff et~al\mbox{.}}{2013}]%
        {wltdg-gsdeud-2013}
{K.~L. Wagstaff}, {N.~L. Lanza}, {D.~R. Thompson}, {T.~G. Dietterich}, {and}
  {M.~S. Gilmore}. 2013.
\newblock \showarticletitle{Guiding Scientific Discovery with Explanations
  using {DEMUD}}. In {\em Proceedings of the Association for the Advancement of
  Artificial Intelligence AAAI 2013 Conference}.
\newblock


\bibitem[\protect\citeauthoryear{Xiao, Wang, Zhang, and Xu}{Xiao
  et~al\mbox{.}}{2014}]%
        {xiao2014}
{Yingchao Xiao}, {Huangang Wang}, {Lin Zhang}, {and} {Wenli Xu}. 2014.
\newblock \showarticletitle{Two methods of selecting Gaussian kernel parameters
  for one-class SVM and their application to fault detection}.
\newblock {\em Knowledge-Based Systems\/}  {59} (2014), 75--84.
\newblock


\bibitem[\protect\citeauthoryear{Xue, Yan, Roddy, and Varma}{Xue
  et~al\mbox{.}}{2006}]%
        {Xue:06}
{Feng Xue}, {Weizhong Yan}, {Nicholas Roddy}, {and} {Anil Varma}. 2006.
\newblock \showarticletitle{Operational Data Based Anomaly Detection for
  Locomotive Diagnostics}. In {\em International Conference on Machine
  Learning}. 236--241.
\newblock


\bibitem[\protect\citeauthoryear{Zhang, Sconyers, Byington, Patrick, Orchard,
  and Vachtsevanos}{Zhang et~al\mbox{.}}{2008}]%
        {Zhang:08}
{Bin Zhang}, {C. Sconyers}, {C. Byington}, {R. Patrick}, {M. Orchard}, {and}
  {G. Vachtsevanos}. 2008.
\newblock \showarticletitle{Anomaly detection: A robust approach to detection
  of unanticipated faults}. In {\em Prognostics and Health Management, 2008.
  PHM 2008. International Conference on}. 1--8.
\newblock
\showDOI{%
\url{http://dx.doi.org/10.1109/PHM.2008.4711445}}


\bibitem[\protect\citeauthoryear{Zhang and Zulkernine}{Zhang and
  Zulkernine}{2006}]%
        {zhang06}
{Jiong Zhang} {and} {Mohammad Zulkernine}. 2006.
\newblock \showarticletitle{Anomaly Based Network Intrusion Detection with
  Unsupervised Outlier Detection}. In {\em IEEE International Conference on
  Communications}, Vol.~5. IEEE, 2388--2393.
\newblock


\bibitem[\protect\citeauthoryear{Zhu and Hastie}{Zhu and Hastie}{2001}]%
        {klr2}
{Ji Zhu} {and} {Trevor Hastie}. 2001.
\newblock \showarticletitle{Kernel Logistic Regression and the Import Vector
  Machine}. In {\em Journal of Computational and Graphical Statistics}. MIT
  Press, 1081--1088.
\newblock


\end{thebibliography}

\appendix

\section{Comparison to Our Previous Work}\label{appendix:prev}

While the general goals and conclusions of our previously published work are similar to this study there are many key differences in benchmark construction and evaluation of results.

In \cite{odd2013} we identified the problem dimensions of \textbf{relative frequency},\textbf{point difficulty} and \textbf{clusteredness} and created a corpus of benchmarks varying these properties across several values, using the same 19 mothersets as in this study.

In \cite{bencon2015} we added the problem dimension of \textbf{Feature Irrelevance}.

These studies used a different battery of algorithms than the current study; we believe this study uses a more representative set of algorithms for evaluation. This study also adds a control group setting for each problem dimension, including the addition of the synthetic motherset. While the previous studies may have made a convincing argument for the impact of the problem dimensions, this study explicitly demonstrates that manipulating these problem dimensions has a statistically significant impact in most cases.

In previous studies, evaluation was done with a linear regression model, observing the coefficients of various factors as evidence of our claims. Similar methods are applied here, but we compare the performance of multiple models to strengthen these claims.

In this study the corpus of benchmarks itself is also evaluated more rigorously. We perform a statistical hypothesis test on each algorithm's output to each benchmark. Benchmarks for which every algorithm fails to differentiate itself from a random ranking are discarded as unsuitable. Conversely, a trivial algorithm is introduced to help identify benchmarks which might be too trivially easy.

\section{The Synthetic Motherset}\label{appendix:synth}

The synthetic motherset was generated by producing 10,000 candidate normals and 10,000 candidate anomalies from two different multivariate distributions with the intention of being able to manipulate all problem dimensions with ease. The candidate normals are drawn from a multivariate gaussian with a covariance matrix of $I$; that is, each feature is drawn from the standard normal distribution independently of the others. The anomalies are drawn uniformly from the hyper-cube defined by the range $(-4,4)$ in each dimension. Both distributions have ten dimensions; that is, each point exists in $R^{10}$.

\section{Choosing a Confusing Partition of Classes}\label{appendix:confuse}

Our heuristic procedure begins by training a Random Forest as in \cite{Breiman} to solve the multi-class classification problem.  Then we calculate the amount of confusion between each pair of classes. For each data point $x_i$, the Random Forest computes an estimate of $P(\hat{y_i}=k|x_i)$, the predicted probability that $x_i$ belongs to class $k$.  We construct a confusion matrix $C$ in which cell $C_{j,k}$ contains the sum of $P(\hat{y_i}=k|x_i)$ for all $x_i$ whose true class $y_i = j$.  We then define a graph in which each node is a class and each edge (between two classes $j$ and $k$) has a weight equal to $C[j,k]+C[k,j]$.  This is the (un-normalized) probability that a data point in class $j$ will be confused with a data point in class $k$ or vice versa.  We then compute the maximum weight spanning tree of this (complete) graph to identify a graph of ``most-confusable'' relationships between pairs of classes.  We then two-color this tree so that no adjacent nodes have the same color. The two colors define the two classes of points.

This approximately maximizes the confusion between the candidate normal and candidate anomaly data points and also tends to make both classes diverse, which increases semantic variation in both sets.

\section{Parameterizing the Battery of Algorithms}\label{appendix:parameters}

\subsection{RKDE}
We implemented the approach described by \cite{kim:08}. We employed a Gaussian radial basis kernel, with kernel bandwidth selected by the Jakkula heuristic. We optimized over a Hampel loss function with the additional parameters set as suggested by the authors.

\subsection{EGMM}
To reduce the computational cost of fitting, and to improve the numerical stability of the process, we first transformed each benchmark dataset via principle component analysis. We selected principle components (in descending eigenvalue order) to retain 95\% of the variance.
To generate the members of the ensemble, we varied the number of clusters $k$ by trying all values in $\{1,2,3,4, 5, 6\}$.  For each value of $k$, we generated 15 GMMs by training on 15 bootstrap replicates of the data and by randomly initializing each replicate.  We then computed the average out-of-bag log likelihood for each value of $k$ and discarded $k$ values whose average log likelihood was less than 0.85 times the average log likelihood of the best value of $k$.  The purpose of this was to discard GMMs that do not fit the data very well. Finally, an anomaly score is computed for each point $x$ by computing the average ``surprise'', which is the average negative log probability density $\frac{1}{L} \sum_{\ell=1}^L -\log P_{\ell}(x)$, where $L$ is the number of fitted GMMs and $P_{\ell}(x)$ is the density assigned by GMM $\ell$ to data point $x$. 
We found in preliminary experiments that this worked better than using the mean probability density $\frac{1}{L} \sum_{\ell=1}^L P_{\ell}(x)$.

\subsection{OCSVM}
We employed the \texttt{libsvm} implementation of Chang and Lin \cite{chang:2011} available at \url{http://www.csie.ntu.edu.tw/~cjlin/libsvm/}. For each benchmark, we employed a Gaussian radial basis function kernel. Selection of kernel bandwidth was done using the DFN method proposed in \cite{xiao2014}, while the parameter $\nu$ was set to 0.03.

\subsection{SVDD}
We implemented this algorithm using the simplex method. We employed a Gaussian radial basis function kernel with kernel bandwidth selection as in \cite{kakde2016}. To generate the needed statistics we tried 100 different kernel bandwidths.

\subsection{LOF}
We employed the \texttt{R} package \texttt{Rlof} available at \url{http://cran.open-source-solution.org/web/packages/Rlof/}. We chose $k$ to be 3\% of the dataset.  This was the smallest value for which LOF would reliably run on all datasets.

\subsection{ABOD}
As described in the main body of the paper we used a KNN approximation of the original algorithm. The only parameter in such an implementation is the choice of $k$, which we set at 0.005 of the data.

\subsection{IFOR}
We used the implementation provided by the authors of \cite{Liu:08}. For our study we found a subsample size of 2048 (or the entire dataset when a benchmark is smaller than this) to work best instead of the 256 suggested by the authors.

\subsection{LODA}
We implemented the algorithm as suggest by the authors of \cite{loda}. Each projection used approximately $\sqrt{d}$ features and a total of $3d$ projections were used, where $d$ is the number of features in the benchmark.

\section{Treating AUC and AP as Random Variables}\label{appendix:random}

The AUC or AP of a random ranking can be seen as a discrete parametric distribution with parameters $n_{\text{norm}}$ (number of normals) and $n_{\text{anom}}$ (number of anomalies). The distribution is parametric because there are ``only'' $(n_{\text{norm}}+n_{\text{anom}})!$ possible rankings, meaning there are a finite number of possible AUC or AP scores for a given set of parameters.

In both cases it is possible to enumerate these scores and compute how much probability mass each score carries, and thus quantiles of these distributions can be computed. However for larger values of $n$ this becomes computationally inefficient.

Instead we computed the quantiles of interest empirically. For each set of parameters present in our corpus, we produced 1 million random ranking samples, computed the AUC and AP of each, and from this estimated the quantiles of interest.


\end{document}